\documentclass[]{bytedance_seed}

\usepackage[toc,page,header]{appendix}

\usepackage[utf8]{inputenc} % allow utf-8 input
\usepackage[T1]{fontenc}    % use 8-bit T1 fonts
\usepackage{hyperref}       % hyperlinks
\usepackage{url}            % simple URL typesetting
\usepackage{booktabs}       % professional-quality tables
\usepackage{amsfonts}       % blackboard math symbols
\usepackage{nicefrac}       % compact symbols for 1/2, etc.
\usepackage{microtype}      % microtypography
\usepackage[dvipsnames,table]{xcolor}         % colors
\usepackage{graphicx}
\usepackage{enumitem}
\usepackage{makecell}
\usepackage{amssymb}
\usepackage{caption}
\usepackage{subcaption}
\usepackage{bbding}
\usepackage{booktabs}
\usepackage{multirow}
\usepackage{tabularx}
\usepackage{array}
\usepackage{amsmath}
\usepackage{wrapfig}
\usepackage{fvextra}
\usepackage{arydshln}

\definecolor{BPpromptblue}{HTML}{1F6388}

\newtcolorbox{BPpromptbox}[1]{
  enhanced,
  breakable,
  width=\linewidth,
  colback=white,
  colframe=BPpromptblue,
  colbacktitle=BPpromptblue,
  coltitle=white,
  boxrule=0.9pt,
  arc=3pt,
  outer arc=3pt,
  boxsep=0pt,
  left=7pt,
  right=7pt,
  top=6pt,
  bottom=6pt,
  toptitle=5pt,
  bottomtitle=5pt,
  fonttitle=\normalfont\rmfamily\small\bfseries,
  title={#1},
  title after break={#1\space (continued)},
  before skip=8pt,
  after skip=8pt
}

\DefineVerbatimEnvironment{BPprompttext}{Verbatim}{
  fontsize=\footnotesize,
  fontfamily=tt,
  breaklines=true,
  breakanywhere=true,
  breaksymbolleft={},
  breaksymbolright={},
  xleftmargin=0pt,
  xrightmargin=0pt,
  tabsize=2
}
\definecolor{BPexamplebrown}{HTML}{9A5A32}

\newtcolorbox{BPexamplebox}[1]{
  enhanced,
  breakable,
  width=\linewidth,
  colback=white,
  colframe=BPexamplebrown,
  colbacktitle=BPexamplebrown,
  coltitle=white,
  boxrule=0.9pt,
  arc=3pt,
  outer arc=3pt,
  boxsep=0pt,
  left=7pt,
  right=7pt,
  top=6pt,
  bottom=6pt,
  toptitle=5pt,
  bottomtitle=5pt,
  fonttitle=\normalfont\rmfamily\small\bfseries,
  title={#1},
  title after break={#1\space (continued)},
  before skip=8pt,
  after skip=8pt
}
\definecolor{BPrGreen}{HTML}{E1F2DA}
\definecolor{BPrPink}{HTML}{FBE1E3}
\definecolor{BPrGreenLine}{HTML}{6A8260}
\definecolor{BPrPinkLine}{HTML}{B77C88}

\tcbset{
  BProllout/user/.style={
    colback=BPrPink!35!white,
    colbacktitle=BPrPink,
    colframe=BPrPinkLine
  },
  BProllout/assistant/.style={
    colback=BPrGreen!30!white,
    colbacktitle=BPrGreen,
    colframe=BPrGreenLine
  },
  BProllout/tool/.style={
    colback=white,
    colbacktitle=BPrGreen,
    colframe=BPrGreenLine
  },
  BProllout/private/.style={
    colback=BPrPink!20!white,
    colbacktitle=BPrPink,
    frame hidden,
    borderline={0.7pt}{0pt}{BPrPinkLine,dashed}
  }
}

\newtcolorbox{BProlloutbox}[2][]{
  enhanced,
  breakable,
  width=\linewidth,
  colback=white,
  colbacktitle=BPrGreen,
  colframe=BPrGreenLine,
  coltitle=black,
  coltext=black,
  boxrule=0.6pt,
  arc=2pt,
  boxsep=0pt,
  left=7pt,
  right=7pt,
  top=5pt,
  bottom=5pt,
  toptitle=3pt,
  bottomtitle=3pt,
  fonttitle=\normalfont\small\bfseries,
  fontupper=\normalfont\small,
  before upper={\setlength{\parindent}{0pt}\setlength{\parskip}{4pt}\raggedright},
  title={#2},
  title after break={#2\space (continued)},
  before skip=6pt,
  after skip=6pt,
  #1
}

\DefineVerbatimEnvironment{BProlloutcode}{Verbatim}{
  fontsize=\footnotesize,
  fontfamily=tt,
  breaklines=true,
  breakanywhere=true,
  breaksymbolleft={},
  breaksymbolright={},
  xleftmargin=0pt,
  xrightmargin=0pt,
  tabsize=2
}

\newlength\savewidth

\renewcommand{\paragraph}[1]{\vspace{1.25mm}\noindent\textbf{#1}}

\newcolumntype{x}[1]{>{\centering\arraybackslash}p{#1pt}}
\newcolumntype{y}[1]{>{\raggedright\arraybackslash}p{#1pt}}
\newcolumntype{z}[1]{>{\raggedleft\arraybackslash}p{#1pt}}

\newcommand{\app}{\raise.17ex\hbox{$\scriptstyle\sim$}}

\definecolor{deemph}{gray}{0.6}

\definecolor{baselinecolor}{gray}{.9}

\usepackage[linesnumbered,ruled,vlined,algo2e]{algorithm2e}
\usepackage{amsmath}
\usepackage{xcolor}

\SetCommentSty{mycommfont}

\title{BehaviorPilot: Behavior-Guided Data Synthesis for Tool-Using Agents}

\author[1, 2, *]{Yuhao Huang}
\author[1, *]{Zengran Wang}
\author[1, \dagger]{Zhishan Zhou}
\author[1]{Yang Zhang}
\author[1, \ddagger]{Weibo Gong}

\affiliation[1]{PICO}
\affiliation[2]{Peking University}

\contribution[*]{Equal Contribution}
\contribution[\dagger]{Project Lead}
\contribution[\ddagger]{Corresponding authors}

\abstract{
Training tool-using agents requires diverse interaction data and reliable execution feedback. Recent approaches adopt scenario-centric synthesis, generating environments and tasks across domains and tool configurations. However, scenario diversity does not ensure broad behavioral coverage: tasks from different scenarios may repeatedly exercise the same interaction patterns while leaving others underrepresented. To enable explicit control over behavioral coverage, we introduce tool-use behaviors as a representation of recurring interaction patterns and propose \textbf{BehaviorPilot}, a behavior-guided data synthesis framework. BehaviorPilot integrates three modules: behavior taxonomy induction to build a hierarchical behavioral representation from open-source trajectories for task synthesis and model diagnosis; behavior-conditioned data synthesis to generate executable tasks targeting selected behavior combinations and retain rollouts that pass task verification and cover the target behaviors; and iterative behavior-guided refinement to target model weaknesses by allocating additional synthesis budget in proportion to per-behavior failure rates on a fixed, behavior-balanced validation set. Applied to Qwen3-8B, BehaviorPilot achieves the best benchmark-average scores among the evaluated baselines on VitaBench, BFCL V4 multi-turn, and \ensuremath{\tau^2}-Bench. Notably, the resulting 8B model surpasses Qwen3-32B in overall performance. Data and models will be made publicly available.
}

\date{\today}
\correspondence{Weibo Gong (gongweibo@bytedance.com)}

\begin{document}
\maketitle

%不需要目录就注释掉 注意目录不要和第一页放在一块 要有\newpage
%\newpage
%\tableofcontents
%\newpage

% \input{sections/abstract}
\section{Introduction}

Large language model (LLM) agents interact with external systems through tools to retrieve information, execute operations, and modify environment states~\citep{yao2022react,schick2023toolformer}. Effective tool use requires more than selecting a function: agents must construct valid arguments, propagate information across calls, satisfy constraints, and adapt to intermediate observations~\citep{li2023apibank,lu2025toolsandbox,yao2025taubench}. We propose to explicitly represent these recurring interaction patterns as tool-use behaviors. Learning to apply and combine them is key to reliable tool-using agents~\citep{yu2025demystifying,zhang2026nemotron}.

Agents learn these interaction behaviors through supervised fine-tuning (SFT) on tool-use trajectories~\citep{qin2024toolllm,zeng2024agenttuning} and reinforcement learning (RL) with task-level feedback~\citep{xi2025agentgymrl,wang2026awm}. The effectiveness of both approaches depends on the quality and diversity of the interaction experiences represented in training data, as well as the reliability of task feedback~\citep{liu2024apigen,prabhakar2025apigen}. To provide such experiences at scale, a common approach is scenario-centric synthesis, which constructs scenarios across different domains and tool configurations~\citep{song2026envscaler,hu2025agentgen}. Within each scenario, tasks are generated from the available tools, data, and constraints, producing training experiences shaped by the selected environments and task designs~\citep{dong2026agentworld}.

\begin{figure*}[t]
    \centering
    \includegraphics[width=\textwidth]{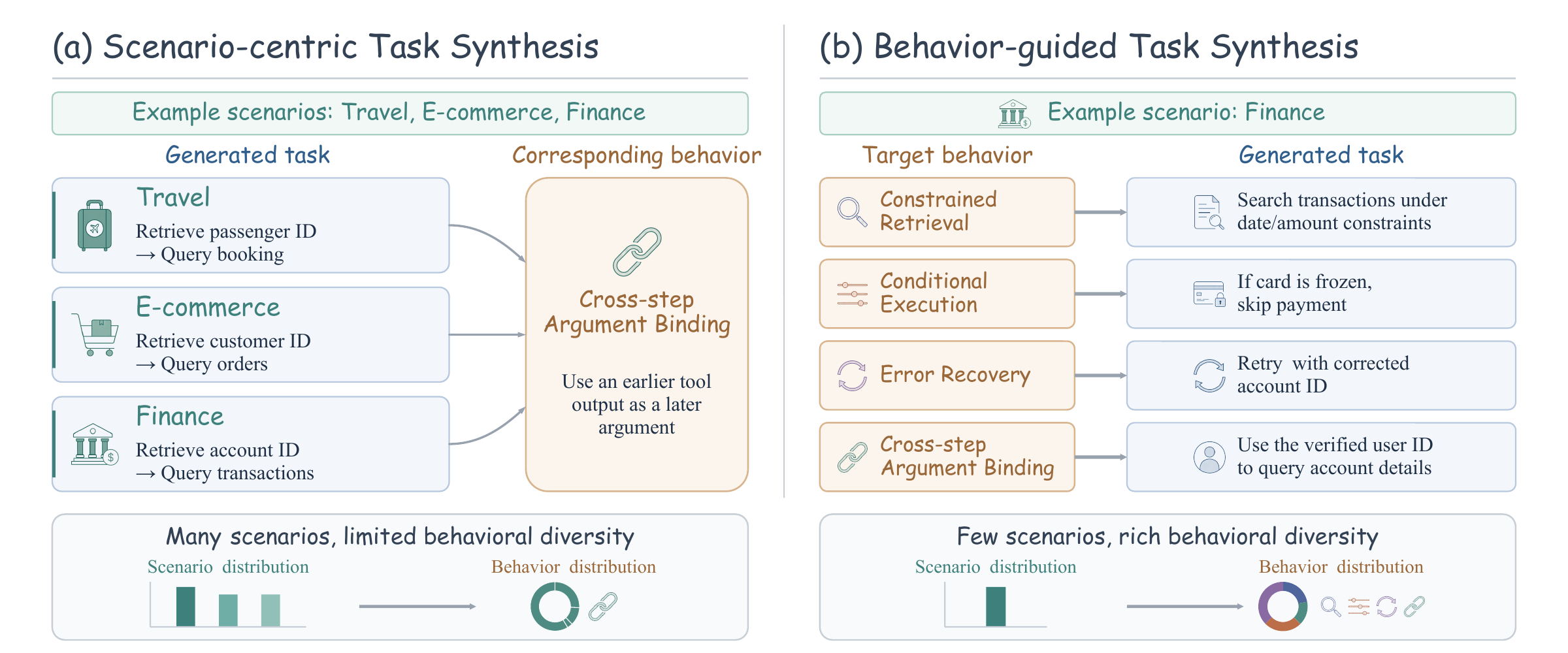}
    \caption{
    Scenario-centric versus behavior-guided synthesis.
    (a) Tasks across distinct scenarios may invoke the same behavior, but cover only a narrow set of behaviors.
    (b) A single executable environment can instead target diverse behaviors, yielding balanced behavior coverage.
    }
    \label{fig:behavior_insight}
\end{figure*}

However, scenario-centric task synthesis does not necessarily ensure broad coverage of tool-use behaviors. As illustrated in Figure~\ref{fig:behavior_insight}, tasks from different domains may still share the same interaction pattern: retrieving a passenger identifier before querying a booking and retrieving a customer identifier before querying an order both involve cross-step argument binding. Conversely, a single environment can support distinct behaviors, including constrained retrieval and error recovery. Scenario diversity alone therefore does not fully characterize the behavioral composition of training data. Without an explicit behavioral representation, it is difficult to assess which patterns the data covers and decide which tasks to generate when a model remains weak.

To make behavioral coverage explicit and target model weaknesses, we propose \textbf{BehaviorPilot}, a behavior-guided data synthesis framework for tool-using agents. BehaviorPilot first induces a hierarchical taxonomy of recurring tool-use behaviors from open-source trajectories, providing a shared representation for both task synthesis and model diagnosis. It then generates behavior-conditioned executable tasks and retains rollouts that pass task verification and cover the target behaviors. The verified trajectories support supervised fine-tuning, while the executable tasks and verifiers support reinforcement learning. During iterative refinement, BehaviorPilot allocates additional synthesis budget in proportion to per-behavior failure rates measured on a fixed, behavior-balanced validation set. Applied to Qwen3-8B, BehaviorPilot outperforms all evaluated baselines on the average scores of VitaBench, BFCL V4 multi-turn, and \begin{math}\tau^2\end{math}-Bench. It achieves an overall score of 40.6, exceeding Qwen3-8B and Qwen3-32B by 12.8 and 4.7 points, respectively.

Our contributions are as follows:
\begin{itemize}
    \item We introduce tool-use behaviors as explicit data-synthesis targets and induce a hierarchical taxonomy from tool-use trajectories, addressing limited behavioral-coverage control in scenario-centric synthesis.
    \item We propose \textbf{BehaviorPilot}, which integrates behavior taxonomy induction, behavior-conditioned data synthesis, and iterative behavior-guided refinement to translate behavioral targets into executable tasks and refine training data based on per-behavior validation.
    \item Extensive experiments on VitaBench, BFCL V4 multi-turn, and \ensuremath{\tau^2}-Bench show that BehaviorPilot, applied to Qwen3-8B, achieves the best benchmark-average scores among the evaluated baselines and outperforms Qwen3-32B overall.
\end{itemize}

\section{Related work}

\subsection{Tool-Use Data Synthesis for Agent Learning}
Training tool-using agents requires data connecting user requests, tool calls, and observations. Existing methods synthesize tasks and collect trajectories at scale using different execution and feedback sources. We group these methods into three approaches: real-tool-based synthesis~\citep{qin2024toolllm,xu2025toucan}, LLM-simulated synthesis~\citep{liu2025toolace,xu2026envace}, and synthetic-environment-based synthesis~\citep{song2026envscaler,tu2026scaleenv}.

Real-tool-based methods construct tasks and collect trajectories using existing tools and services~\citep{patil2024gorilla,prabhakar2025apigen,liu2024apigen}. ToolLLM~\citep{qin2024toolllm} generates instructions over real-world APIs and searches for solution paths, while TOUCAN~\citep{xu2025toucan} collects validated trajectories through tool execution in MCP environments. However, paid access, request limits, and tool unavailability make scaling costly and slow.

LLM-simulated methods use language models to generate interactions without executing the underlying tools. ToolACE~\citep{liu2025toolace} synthesizes tool-use dialogues with simulated responses, while EnvACE~\citep{xu2026envace} uses a shared policy to alternate between taking actions and rehearsing environment responses. These methods reduce dependence on external services, but simulated responses may not reflect actual tool behavior or state changes.

Synthetic-environment-based methods construct executable environments with explicit tool interfaces, persistent states such as databases, and task-specific verifiers~\citep{song2026envscaler,tu2026scaleenv,wang2026awm,xu2026envfactory,dong2026agentworld}. Tool calls execute code to read or update the environment state and return observations, while verifiers check whether task requirements are satisfied. This supports trajectory collection and outcome evaluation without relying on live services or LLM-simulated tool responses. BehaviorPilot follows this approach, using explicit behavioral targets to guide task generation within these environments.

\subsection{Task Generation in Executable Environments}

Building on executable environments, recent work studies how to scale environment construction and generate tasks that support agent learning.

Several methods emphasize scalable environment and task construction. EnvScaler~\citep{song2026envscaler} generates environment skeletons and task scenarios with rule-based validators. ScaleEnv~\citep{tu2026scaleenv} combines procedural testing with tool-dependency graph expansion to construct executable environments and verifiable tasks. AWM~\citep{wang2026awm} generates database-backed environments with executable tools and task-specific success criteria. Agent-World~\citep{dong2026agentworld} broadens domain coverage through environment discovery and generates verifiable tasks within the resulting environments. Agent diagnosis then guides further environment and task expansion. EnvFactory~\citep{xu2026envfactory} grounds environment construction in online resources and guides trajectory generation with tool-interaction graphs.

Other work couples environment synthesis with multi-stage or adaptive agent training. AgentScaler~\citep{fang2026agentscaler} constructs heterogeneous executable environments and trains agents in two stages, progressing from general tool use to domain specialization. SPADE~\citep{liu2026spade} jointly trains an environment designer and a reasoning agent through self-play, using hint-based regret to adapt the difficulty of generated environments.

These methods generate more varied environments and tasks or adjust task difficulty based on agent performance. However, different tasks can still exercise the same tool-use patterns, while some behaviors receive little training. In contrast, BehaviorPilot uses behavioral targets to guide generation, enabling explicit control over the types and distribution of behaviors in the generated tasks.

\section{Methods}

\subsection{Problem Formulation}

An executable environment \begin{math}\mathcal{E}\end{math} provides an underlying data state and a set of callable tools. Tool calls operate on this state and return observations according to predefined execution logic. Let \begin{math}\mathcal{B}=\{b_1,\ldots,b_K\}\end{math} denote a set of behavior groups, each describing a recurring tool-use pattern. We formulate behavior-guided data synthesis as constructing a task in \begin{math}\mathcal{E}\end{math} conditioned on a target behavior combination \begin{math}\mathbf{b}\subseteq\mathcal{B}\end{math}. Each task is defined by an initial state \begin{math}s_0\end{math}, a user request \begin{math}u\end{math}, and a task-specific verifier. Interaction yields a trajectory
\begin{equation}
\tau=(u,a_1,o_1,\ldots,a_T,o_T),
\end{equation}
where \begin{math}a_t\end{math} denotes a tool call and \begin{math}o_t\end{math} denotes the resulting observation. Each call may query or modify the environment state, and the verifier determines whether the interaction satisfies the task requirements. Verified trajectories provide supervised interaction data; the tasks, environments, and verifiers support RL.
% The objective is to use behavioral targets to guide data construction and improve agents' generalization to unseen tool-use tasks.

\subsection{Framework Overview}

\begin{figure*}[t]
    \centering
    \includegraphics[width=\textwidth]{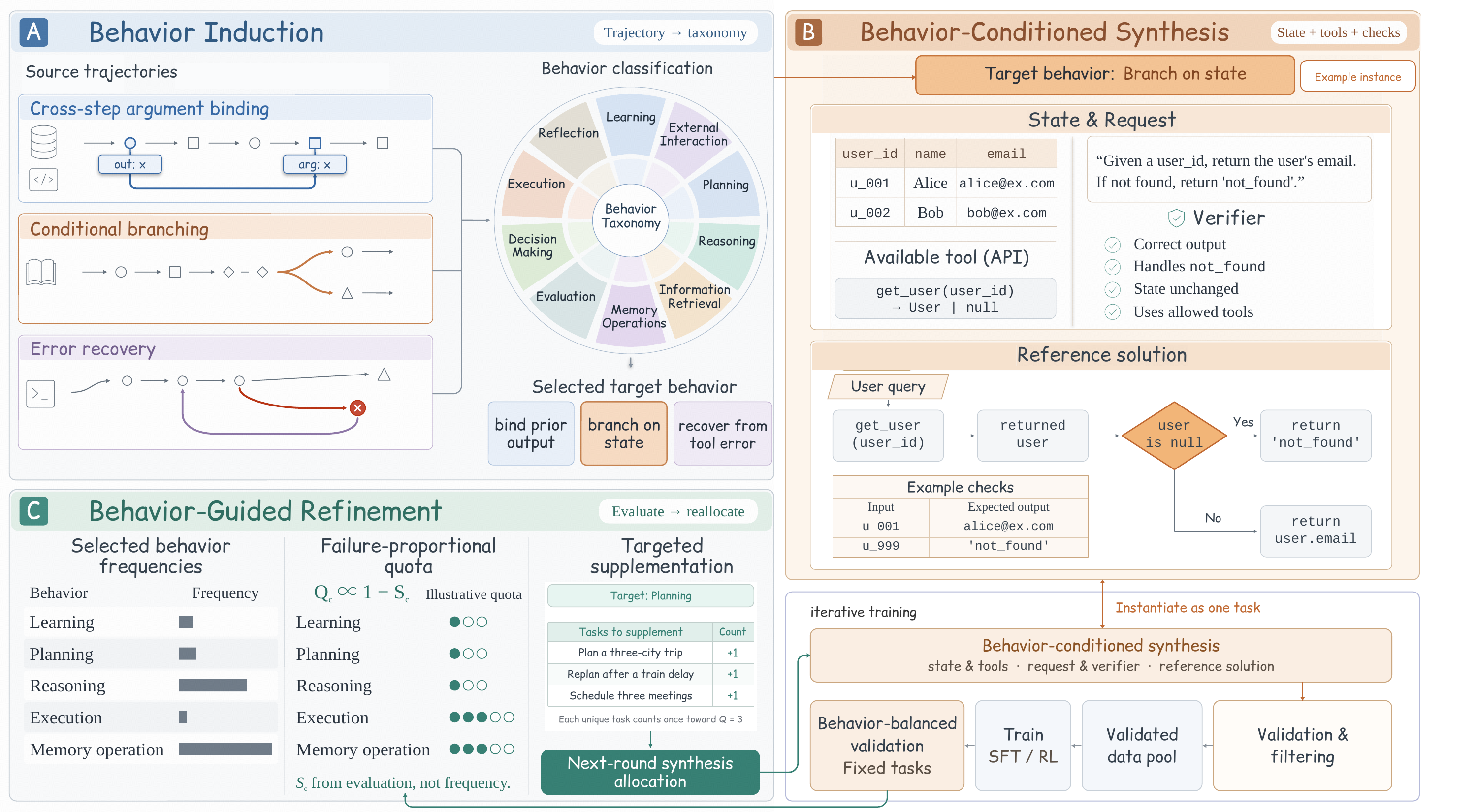}
    \caption{
        Overview of BehaviorPilot. (A) Inducing a behavior taxonomy
    from tool-use trajectories. (B) Synthesizing behavior-conditioned
    tasks and verifying trajectories for training. (C) Iteratively
    refining the model by allocating synthesis budgets according to
    behavior-level validation failure rates.
    }
    \label{fig:framework}
    % \vspace{-0.2cm}
\end{figure*}

We propose \textbf{BehaviorPilot}, a behavior-guided data synthesis framework that addresses the limited control over behavioral coverage in scenario-centric synthesis. As shown in Figure~\ref{fig:framework}, BehaviorPilot first induces a hierarchical behavior taxonomy from tool-use trajectories, establishing a shared representation for task synthesis and model diagnosis. It then generates executable tasks conditioned on behavior combinations and validates task feasibility and behavioral coverage. Agent rollouts that pass task verification and cover the target behaviors are retained for SFT, while the executable tasks and verifiers support RL. During refinement, behavior-level evaluation identifies weaknesses and allocates synthesis budgets by failure rate.

\subsection{Behavior Taxonomy Induction}

To establish a shared behavioral representation for data synthesis and model diagnosis, we construct a behavior taxonomy from publicly available tool-use trajectories. For each trajectory \begin{math}\tau_i\end{math}, an LLM examines its tool calls, arguments, and observations to produce descriptions of recurring interaction patterns, such as cross-step argument binding, conditional branching, and error recovery. These descriptions capture tool-use behaviors rather than domain-specific entities or tool names. The induction and consolidation process is summarized as
\begin{equation}
\mathcal{T}_{\mathrm{src}}
\xrightarrow{\operatorname{Induct}}
\mathcal{Y}
\xrightarrow{\operatorname{Cluster}}
\{C_1,\ldots,C_K\}
\xrightarrow{\operatorname{Merge}}
\mathcal{B}=\{b_1,\ldots,b_K\},
\end{equation}
where \begin{math}\mathcal{T}_{\mathrm{src}}\end{math} is the collection of source trajectories, \begin{math}\mathcal{Y}\end{math} is the pool of induced behavior descriptions, each \begin{math}C_k\end{math} clusters semantically similar descriptions, and \begin{math}b_k\end{math} is the resulting Behavior Group. 

We organize \begin{math}\mathcal{B}\end{math} into a three-level Category--Subcategory--Behavior hierarchy. Categories represent broad types of tool-use interactions, subcategories group related behaviors, and leaf-level behaviors specify concrete interaction patterns. Each behavior retains an operational description and representative source trajectories for task synthesis and model diagnosis. Figure~\ref{fig:behavior_taxonomy} illustrates this hierarchy.
% A task may target a combination of behaviors \begin{math}\mathbf{b}\subseteq\mathcal{B}\end{math}, allowing multiple interaction patterns to be incorporated into a single task.

\subsection{Behavior-Conditioned Data Synthesis}

\begin{figure*}[t]
    \centering
    \begin{minipage}[b]{0.56\textwidth}
        \centering
        \includegraphics[width=\linewidth]{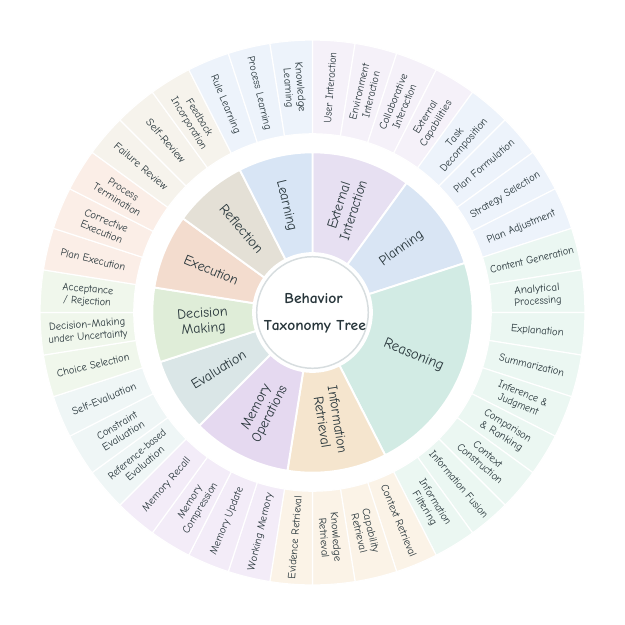}
        \caption{The hierarchical organization of the behavior taxonomy.}
        \label{fig:behavior_taxonomy}
        \label{fig:taxonomy_tree}
    \end{minipage}
    \hfill
    \begin{minipage}[b]{0.36\textwidth}
        \centering
        \includegraphics[width=\linewidth]{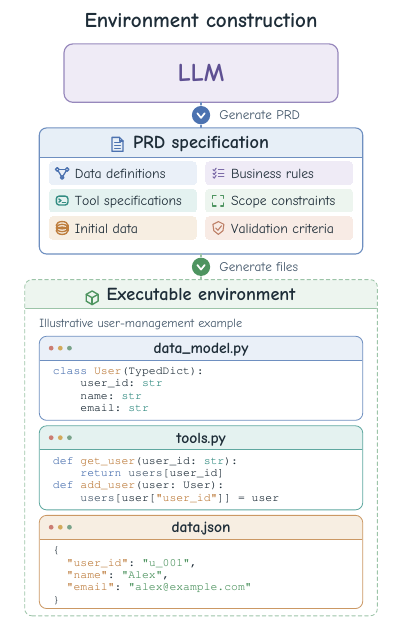}
        \caption{The workflow for generating executable environments.}
        \label{fig:environment_construction}
    \end{minipage}
\end{figure*}

Building on the taxonomy, we use behavior combinations as explicit targets for task synthesis. To support their instantiation, we construct executable environments from LLM-generated business specifications, which guide the generation of database schemas, executable tools, and initial data, as illustrated in Figure~\ref{fig:environment_construction}. Within each environment, an LLM-based planner selects a combination \begin{math}\mathbf{b}\subseteq\mathcal{B}\end{math} supported by its tools and state transitions. It then designs a task whose goals, constraints, and tool dependencies are intended to elicit these behaviors. The environment provides the business context and executable operations, while the selected behaviors guide what the task requires the agent to do.

% The task specification is instantiated as a user request, a reference solution, and a task-specific verifier:
% \begin{equation}
% (u,r,v)=\operatorname{Generate}(\mathcal{E},\mathbf{b},q),
% \end{equation}
% where \begin{math}q\end{math} denotes the planned task specification, and \begin{math}u\end{math}, \begin{math}r\end{math}, and \begin{math}v\end{math} denote the user request, reference solution, and verifier, respectively. To verify behavioral coverage, an LLM identifies the behaviors involved in the reference solution and checks whether they include all target behaviors in \begin{math}\mathbf{b}\end{math}.
% Formally, this requires
% \begin{equation}
% \mathbf{b}\subseteq\widehat{\mathcal{B}}(r),
% \qquad
% \widehat{\mathcal{B}}(r)=\operatorname{Identify}(r),
% \end{equation}
% where \begin{math}\widehat{\mathcal{B}}(r)\end{math} is the set of behaviors identified in the reference solution. If any target behavior is missing, we revise the task and reference solution and repeat the check. 
% After the behavioral coverage check, we remove duplicate or highly similar tasks and execute their reference solutions to confirm feasibility. We then collect agent rollouts for the validated tasks. An LLM identifies the behaviors in each rollout and checks whether it covers all target behaviors, without requiring the same tool-call sequence as the reference solution. Only rollouts that pass both the task verifier and this trajectory-level behavior-coverage check are retained for SFT. The validated tasks also support RL, with verifiers providing rewards during interaction.
The task specification is instantiated as a user request, a reference solution, and a task-specific verifier:
\begin{equation}
(u,r,v)=\operatorname{Generate}(\mathcal{E},\mathbf{b},q),
\end{equation}
where \begin{math}q\end{math} denotes the planned task specification, and \begin{math}u\end{math}, \begin{math}r\end{math}, and \begin{math}v\end{math} denote the user request, reference solution, and verifier, respectively. To verify behavioral coverage, an LLM identifies the behaviors in the reference solution and checks that all target behaviors in \ensuremath{\mathbf{b}} are included. If any target behavior is missing, we revise the task and reference solution and repeat the check.
After the behavioral coverage check, we remove duplicate or highly similar tasks and execute their reference solutions to confirm feasibility. We then collect agent rollouts for the validated tasks. An LLM identifies the behaviors in each rollout and checks whether it covers all target behaviors, without requiring the same tool-call sequence as the reference solution. Only rollouts that pass both the task verifier and this trajectory-level behavior-coverage check are retained for SFT. The validated tasks also support RL, with verifiers providing rewards during interaction.

\subsection{Iterative Behavior-Guided Refinement}

We further refine the training data by identifying behaviors on which the model remains weak and targeting them in subsequent synthesis rounds. We first construct a fixed, behavior-balanced validation set using the same executable environments as training but a separate set of tasks. Each validation task targets one designated behavior, with equal task counts across groups for comparable behavior-level success rates.

% After each training round \begin{math}t\end{math}, we evaluate the updated model on this \textcolor{red}{set. For behavior} \begin{math}b_k\end{math}, the success rate \textcolor{red}{is}
% \begin{equation}
% s_k^{(t)}
% =
% \frac{1}{N_k}
% \sum_{i=1}^{N_k}
% \mathbf{1}\!\left[v_{ki}(\tau_{ki}^{(t)})=1\right].
% \end{equation}
% where \begin{math}N_k\end{math} is the number of validation tasks targeting \begin{math}b_k\end{math}, \begin{math}\tau_{ki}^{(t)}\end{math} is the rollout produced on the \begin{math}i\end{math}-th such task, and \begin{math}v_{ki}\end{math} is its task verifier. We allocate a fixed synthesis budget across behaviors \textcolor{red}{by} failure \textcolor{red}{rate:}
% \begin{equation}
% Q_k^{(t+1)}
% =
% B_t\cdot
% \frac{1-s_k^{(t)}}
% {\sum_{j=1}^{K}(1-s_j^{(t)})},
% \end{equation}
% where \begin{math}B_t\end{math} is the total synthesis budget for the next round and \begin{math}Q_k^{(t+1)}\end{math} is the number of new training tasks allocated to behavior \begin{math}b_k\end{math}. Behaviors with lower success rates receive more tasks. Each \textcolor{red}{task} counts toward one primary behavior's quota, even \textcolor{red}{if} it involves multiple behaviors. The synthesis pipeline receives only aggregate behavior-level statistics, not validation tasks or solutions, limiting direct \textcolor{red}{validation-set} adaptation.
After each training round \begin{math}t\end{math}, we evaluate the updated model on this set. For each behavior \ensuremath{b_k}, we record its validation success rate \ensuremath{s_k^{(t)}}, defined as the fraction of its validation tasks completed successfully in round \ensuremath{t}. We allocate a fixed synthesis budget across behaviors in proportion to failure rates:
\begin{equation}
Q_k^{(t+1)}
=
B_t\cdot
\frac{1-s_k^{(t)}}
{\sum_{j=1}^{K}(1-s_j^{(t)})},
\end{equation}
where \begin{math}B_t\end{math} is the total synthesis budget for the next round and \begin{math}Q_k^{(t+1)}\end{math} is the number of new training tasks allocated to behavior \begin{math}b_k\end{math}.

We repeat evaluation, targeted synthesis, and training until predefined behavior-level targets or budget limits are met. Public benchmarks are reserved for final generalization evaluation and do not guide synthesis or stopping decisions.

\section{Experiments}

\subsection{Datasets and Evaluation Metrics}
\label{subsec:datasets}

Following prior work on training tool-using agents~\citep{song2026envscaler,tu2026scaleenv,wang2026awm}, we evaluate on three public agentic tool-use benchmarks. 
\textbf{BFCL V4}~\citep{patil2025bfcl} evaluates function-calling capabilities across diverse tool schemas and parameter configurations. We use its Multi-Turn subset, which tests whether agents can maintain context, select appropriate tools across turns, and handle cases where required functions or parameters are not explicitly provided. It consists of Base, Long Context, Missing Function, and Missing Parameter tasks. \textbf{\begin{math}\tau^2\end{math}-Bench}~\citep{barres2025tau2} evaluates conversational agents in stateful environments in which tool calls can query and update persistent records. We report task-level success on its Retail, Telecom, and Airline domains. \textbf{VitaBench}~\citep{he2026vitabench} covers multi-step and multi-turn tool use in life-service scenarios, requiring agents to integrate information across tools and satisfy user goals within realistic business workflows. We report task-completion results on its Cross, Delivery, In-store, and OTA subsets. Results are reported as task success rates averaged over four runs.

\subsection{Implementation Details}
\label{subsec:implementation}

% For behavior taxonomy induction, we use DeepSeek-V4-Pro-Preview~\citep{deepseekai2026deepseekv4} to extract and consolidate behavior descriptions from source trajectories. The resulting taxonomy comprises 10 categories, 40 subcategories, and 100 leaf-level behaviors.

% For behavior-conditioned data synthesis, we also use DeepSeek-V4-Pro-Preview to construct 100 executable environments and generate task--solution--verifier triples conditioned on selected behavior combinations. Qwen3.6-27B~\citep{qwen2026qwen36_27b} serves as the teacher model for trajectory collection, yielding more than 9,000 verified trajectories for SFT.

% We adopt Qwen3-8B~\citep{qwen2025qwen3} as the base model and train it with SFT using MS-SWIFT~\citep{zhao2024msswift}, followed by agentic RL using ROLL~\citep{wang2025roll}. RL employs GRPO~\citep{shao2024deepseekmath} with rewards provided by task verifiers. Both stages use the same synthesized environments, while behavior-level validation guides additional data synthesis during iterative refinement. For computational efficiency, we use SFT only to evaluate model performance in the ablation and analysis experiments.

For behavior taxonomy induction, we use DeepSeek-V4-Pro-Preview~\citep{deepseekai2026deepseekv4} to extract and consolidate behavior descriptions from source trajectories. The resulting taxonomy comprises 10 categories, 40 subcategories, and 100 leaf-level behaviors.

For behavior-conditioned data synthesis, we also use DeepSeek-V4-Pro-Preview to construct 100 executable environments and generate task--solution--verifier triples conditioned on selected behavior combinations. Qwen3.6-27B~\citep{qwen2026qwen36_27b} serves as the teacher model for trajectory collection, yielding more than 9,000 verified trajectories for SFT.

We adopt Qwen3-8B~\citep{qwen2025qwen3} as the base model and perform SFT using MS-SWIFT~\citep{zhao2024msswift} on 9,000+ verified trajectories collected across the initial synthesis and three refinement rounds. Using the resulting SFT checkpoint, we collect four rollouts per SFT training task and retain tasks solved in some but not all attempts. We then initialize agentic RL from this same checkpoint and train on fresh online rollouts from the selected tasks using ROLL~\citep{wang2025roll}. RL employs GRPO~\citep{shao2024deepseekmath}, with rewards provided by task verifiers. Both stages use the same synthesized environments, while behavior-level validation guides additional data synthesis during iterative refinement. The validation set covers 100 behaviors, with 50 tasks per behavior, totaling 5,000 tasks. 
% For computational efficiency, we use SFT only to evaluate model performance in the ablation and analysis experiments.

\subsection{Main Results}
\label{subsec:main-results}
We compare BehaviorPilot with Qwen3 models~\citep{qwen2025qwen3} at 4B, 8B, and 32B scales, as well as representative tool-use post-trained models, including TOUCAN~\citep{xu2025toucan}, Simulator~\citep{li2025simula}, ScaleEnv~\citep{tu2026scaleenv}, AWM~\citep{wang2026awm}, EnvFactory~\citep{xu2026envfactory}, EnvScaler~\citep{song2026envscaler}, EnvACE~\citep{xu2026envace}, and SPADE~\citep{liu2026spade}. Table~\ref{tab:main} shows that BehaviorPilot achieves the best average score on all three benchmarks, reaching 20.5 on VitaBench, 53.3 on BFCL V4 Multi-Turn, and 48.1 on \ensuremath{\tau^2}-Bench. With an overall score of 40.6, BehaviorPilot ranks first among all evaluated methods. Its consistent lead across the three benchmarks indicates that behavior-guided training improves tool-use capabilities across diverse tasks and interaction settings.

Compared with Qwen3 models, BehaviorPilot delivers substantial gains at the same model scale. It improves over Qwen3-8B by 12.8 points overall (40.6 versus 27.8). More importantly, BehaviorPilot also surpasses the substantially larger Qwen3-32B model by 4.7 points overall. The advantage holds consistently across all three benchmarks, with improvements of 5.7 points on VitaBench, 5.4 points on BFCL V4 Multi-Turn, and 3.2 points on \ensuremath{\tau^2}-Bench. These results demonstrate that behavior-guided post-training can enable an 8B model to outperform a much larger off-the-shelf model on diverse tool-use tasks.

% We further compare BehaviorPilot with tool-use post-trained baselines. EnvScaler-8B and EnvACE-8B are the two strongest prior methods in overall score, achieving 33.0 and 32.8, respectively. BehaviorPilot outperforms EnvScaler-8B by 7.6 points overall, including gains of 4.7 points on VitaBench, 3.1 points on BFCL V4 Multi-Turn, and 15.2 points on \ensuremath{\tau^2}-Bench. It also exceeds EnvACE-8B by 7.8 points overall, with improvements of 4.5, 7.6, and 11.4 points on the three benchmarks, respectively. These results show that BehaviorPilot provides consistent improvements over strong tool-use post-training approaches across different evaluation settings.

We further compare BehaviorPilot with tool-use post-trained baselines. EnvScaler-8B and EnvACE-8B are the two strongest prior methods in overall score, achieving 33.0 and 32.8, respectively. BehaviorPilot outperforms EnvScaler-8B by 7.6 points overall, including gains of 4.7 points on VitaBench, 3.1 points on BFCL V4 Multi-Turn, and 15.2 points on \ensuremath{\tau^2}-Bench. It also exceeds EnvACE-8B by 7.8 points overall, with improvements of 4.5, 7.6, and 11.4 points on the three benchmarks, respectively. These results show that BehaviorPilot consistently improves over strong tool-use post-training approaches across evaluation settings.

\definecolor{groupgray}{RGB}{238,238,238}
\definecolor{groupblue}{RGB}{226,238,252}

\begin{table}[ht]
  \caption{Main results on agentic tool-use benchmarks. Within the Tool-Use Post-Trained Models group, \textbf{bold} and \underline{underlined} values indicate the best and second-best results, respectively.}
  \label{tab:main}
  \centering
  \setlength{\tabcolsep}{2.5pt}
  \renewcommand{\arraystretch}{1.12}

  \resizebox{\textwidth}{!}{%
  \begin{tabular}{@{}l*{15}{c}@{}}
    \toprule
    \multirow[c]{2}{*}[-0.5ex]{\textbf{Method}}
      & \multicolumn{5}{c}{\textbf{VitaBench}}
      & \multicolumn{5}{c}{\textbf{BFCL V4 MT}}
      & \multicolumn{4}{c}{\textbf{\ensuremath{\tau^2}-Bench}}
      & \multirow[c]{2}{*}[-0.5ex]{\textbf{Overall}} \\
    \cmidrule(lr){2-6}
    \cmidrule(lr){7-11}
    \cmidrule(lr){12-15}
      & Cross & Delivery & In-store & OTA & \textbf{Avg.}
      & Base & Long Ctx & Miss Func & Miss Param & \textbf{Avg.}
      & Retail & Telecom & Airline & \textbf{Avg.} \\
    \midrule

    \multicolumn{16}{c}{%
      \cellcolor{groupgray}
      \strut\textbf{Base Models (4B--32B)}} \\

    Qwen3-4B
      & 1.8 & 13.7 & 18.2 & 4.8 & 9.6
      & 51.0 & 39.0 & 34.0 & 28.0 & 38.0
      & 32.7 & 13.2 & 38.0 & 27.9
      & 25.2 \\
    Qwen3-8B
      & 1.2 & 15.5 & 24.8 & 4.0 & 11.4
      & 55.5 & 41.0 & 36.0 & 35.0 & 41.9
      & 41.2 & 20.8 & 28.0 & 30.0
      & 27.8 \\
    Qwen3-32B
      & 5.3 & 27.0 & 22.5 & 4.5 & 14.8
      & 56.0 & 43.0 & 52.5 & 40.0 & 47.9
      & 59.5 & 27.2 & 48.0 & 44.9
      & 35.9 \\

    \midrule
    \multicolumn{16}{c}{%
      \cellcolor{groupgray}
      \strut\textbf{Tool-Use Post-Trained Models (7B--8B)}} \\

    TOUCAN-7B
      & 0.5 & 6.9 & 3.0 & 0.8 & 2.8
      & 34.3 & 27.8 & 13.4 & 21.1 & 24.1
      & 30.0 & 10.1 & 27.0 & 22.4
      & 16.4 \\
    Simulator-8B
      & 0.0 & 1.2 & 4.8 & 1.0 & 1.8
      & 6.4 & 6.0 & 2.0 & 7.0 & 5.3
      & 49.3 & 15.1 & \textbf{51.0} & \underline{38.5}
      & 15.2 \\
    ScaleEnv-8B
      & 3.0 & \textbf{26.3} & 23.8 & 7.0 & 15.0
      & -- & -- & -- & -- & --
      & \underline{50.9} & \underline{27.2} & 37.5 & \underline{38.5}
      & -- \\
    AWM-8B
      & 1.5 & 16.0 & 19.2 & 4.2 & 10.2
      & 50.8 & 42.1 & 41.3 & 36.8 & 42.7
      & 41.3 & 18.2 & 34.0 & 31.2
      & 28.0 \\
    EnvFactory-8B
      & 2.2 & 24.0 & 22.0 & \underline{10.0} & 14.6
      & 57.4 & \underline{46.5} & 43.0 & 29.4 & 44.1
      & 43.9 & 13.2 & \underline{44.0} & 33.7
      & 30.8 \\
    EnvScaler-8B
      & \textbf{6.0} & 20.8 & \underline{29.5} & 7.0 & 15.8
      & \underline{66.0} & \underline{46.5} & \textbf{48.8} & 39.4 & \underline{50.2}
      & 48.0 & 16.2 & 34.5 & 32.9
      & \underline{33.0} \\
    EnvACE-8B
      & \textbf{6.0} & 24.0 & 27.0 & 7.0 & \underline{16.0}
      & 53.1 & 43.8 & \underline{45.1} & \underline{40.9} & 45.7
      & 48.9 & 17.3 & \underline{44.0} & 36.7
      & 32.8 \\
    SPADE-8B
      & 0.0 & 5.0 & 1.0 & 0.0 & 1.5
      & 58.0 & 43.2 & 36.0 & 30.0 & 41.8
      & 37.8 & 21.2 & 29.5 & 29.5
      & 24.3 \\

    \rowcolor{groupblue}
    \textbf{BehaviorPilot-8B}
      & \underline{5.0} & \underline{26.0} & \textbf{35.0} & \textbf{16.0} & \textbf{20.5}
      & \textbf{70.0} & \textbf{56.5} & 42.5 & \textbf{44.0} & \textbf{53.3}
      & \textbf{60.5} & \textbf{43.9} & 40.0 & \textbf{48.1}
      & \textbf{40.6} \\

    \bottomrule
  \end{tabular}%
  }
\end{table}

\begin{table}[ht]
\caption{Performance across behavior-guided refinement rounds on BFCL and \ensuremath{\tau^2}-Bench. Initial uses the initial synthesized data. Tasks denotes the approximate cumulative number of tasks (k = 1{,}000). Bold marks column bests.}
\label{tab:refinement_rounds}
\centering
\footnotesize
\setlength{\tabcolsep}{3pt}
\renewcommand{\arraystretch}{1.05}

\begin{tabular}{@{}lc*{9}{c}@{}}
\toprule
\multirow[c]{2}{*}[-0.5ex]{Setting}
& \multirow[c]{2}{*}[-0.5ex]{Tasks}
& \multicolumn{5}{c}{\textbf{BFCL}}
& \multicolumn{4}{c}{\textbf{\ensuremath{\tau^2}-Bench}} \\
\cmidrule(lr){3-7}
\cmidrule(lr){8-11}
& & Base & Long Ctx & Miss Func & Miss Param & Avg.
& Retail & Telecom & Airline & Avg. \\
\midrule
Initial & \textasciitilde{}4k
& \textbf{68.5} & 44.0 & 37.5 & 44.0 & 48.5
& 59.1 & 33.3 & 42.0 & 44.8 \\
+1 round & \textasciitilde{}6k
& 60.0 & 53.0 & 38.0 & \textbf{45.0} & 49.0
& 60.2 & 27.0 & \textbf{51.9} & 46.4 \\
+2 rounds & \textasciitilde{}8k
& 62.0 & 52.5 & 41.0 & 44.0 & 49.9
& 64.2 & \textbf{44.0} & 34.0 & 47.4 \\
+3 rounds & \textasciitilde{}10k
& 65.0 & \textbf{58.0} & \textbf{46.0} & 40.0 & \textbf{52.3}
& \textbf{65.8} & 42.1 & 44.2 & \textbf{50.7} \\
\bottomrule
\end{tabular}
% \vspace{-0.1cm}
\end{table}

\subsection{Model Performance during Iterative Behavior-Guided Refinement}
\label{subsec:iterative-refinement}
We perform three rounds of behavior-guided refinement, using behavior-level validation to guide the synthesis of approximately 2{,}000 additional tasks per round and updating the model through SFT. This expands the training data from approximately 4{,}000 to 10{,}000 tasks. Table~\ref{tab:refinement_rounds} shows that both benchmark averages improve after every round: BFCL increases from 48.5 to 52.3, and \ensuremath{\tau^2}-Bench from 44.8 to 50.7. BFCL gains are concentrated in long-context and missing-function tasks, whose scores rise from 44.0 to 58.0 and from 37.5 to 46.0, respectively. On \ensuremath{\tau^2}-Bench, all three domains finish above their initial scores, with the largest absolute gain in telecom. The final round achieves the highest average on both benchmarks. Together, these results support iterative behavior-guided supplementation as a practical way to extend the benefits of initial data synthesis and sustain improvements in aggregate tool-use performance.

\begin{table}[t]
\caption{Performance across synthesis strategies, environment counts, and task budgets on BFCL and \ensuremath{\tau^2}-Bench. Behavior-agnostic denotes synthesis without explicit behavioral guidance. Bold marks column bests.}
\label{tab:behavior_scaling}
\centering
\small
\setlength{\tabcolsep}{4pt}
\renewcommand{\arraystretch}{1.10}
\resizebox{0.9\linewidth}{!}{%
\begin{tabular}{@{}lcc*{9}{c}@{}}
\toprule
\multirow[c]{2}{*}[-0.5ex]{Setting}
& \multirow[c]{2}{*}[-0.5ex]{Env.}
& \multirow[c]{2}{*}[-0.5ex]{Tasks}
& \multicolumn{5}{c}{\textbf{BFCL}}
& \multicolumn{4}{c}{\textbf{\ensuremath{\tau^2}-Bench}} \\
\cmidrule(lr){4-8}
\cmidrule(lr){9-12}
& & & Base & Long Ctx & Miss Func & Miss Param & Avg.
& Retail & Telecom & Airline & Avg. \\
\midrule
Qwen3-8B & -- & --
& 55.5 & 41.0 & 36.0 & 35.0 & 41.9
& 41.2 & 20.8 & 28.0 & 30.0 \\
\midrule
\multicolumn{12}{l}{\textit{Joint environment and task scaling}} \\
Behavior-agnostic & 100 & 4k
& 62.7 & 43.1 & 36.2 & 27.4 & 42.4
& 27.2 & \textbf{16.7} & 24.0 & 22.6 \\
Behavior-agnostic & 200 & 8k
& 66.9 & \textbf{46.5} & 40.0 & 27.0 & 45.1
& 32.5 & 14.9 & 20.0 & 22.5 \\
Behavior-agnostic & 300 & 12k
& 67.3 & 46.1 & 39.2 & \textbf{28.4} & 45.3
& 30.7 & 13.2 & \textbf{26.0} & 23.3 \\
Behavior-agnostic & 400 & 16k
& \textbf{67.4} & \textbf{46.5} & \textbf{40.2} & \textbf{28.4} & \textbf{45.6}
& \textbf{36.8} & 14.0 & 22.0 & \textbf{24.3} \\
\midrule
\multicolumn{12}{l}{\textit{Behavior guidance and environment diversity}} \\
Behavior-agnostic & 400 & 4k
& 66.5 & 41.1 & 37.3 & 28.5 & 43.4
& 31.8 & 15.5 & 14.3 & 20.5 \\
\textbf{BehaviorPilot} & 100 & 4k
& \textbf{68.5} & \textbf{44.0} & \textbf{37.5} & \textbf{44.0} & \textbf{48.5}
& \textbf{59.1} & \textbf{33.3} & \textbf{42.0} & \textbf{44.8} \\
\bottomrule
\end{tabular}%
}
\end{table}

\begin{figure}[t]
    \centering
    \begin{subfigure}[t]{0.32\textwidth}
        \centering
        \includegraphics[width=\linewidth]{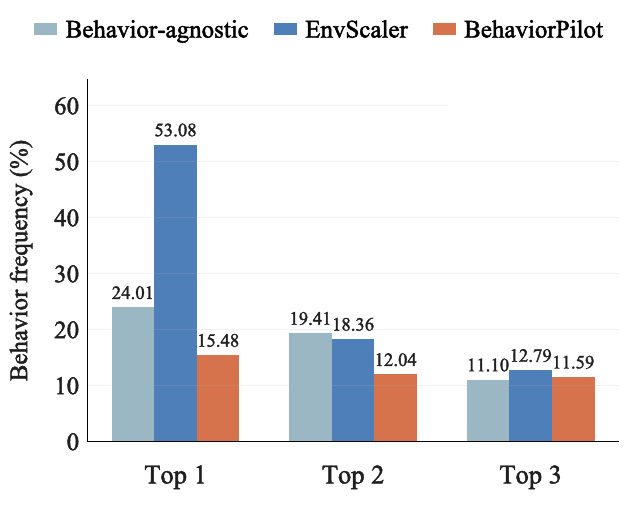}
        \caption{Most frequent behaviors.}
        \label{fig:behavior_distribution_top}
    \end{subfigure}
    \hfill
    \begin{subfigure}[t]{0.32\textwidth}
        \centering
        \includegraphics[width=\linewidth]{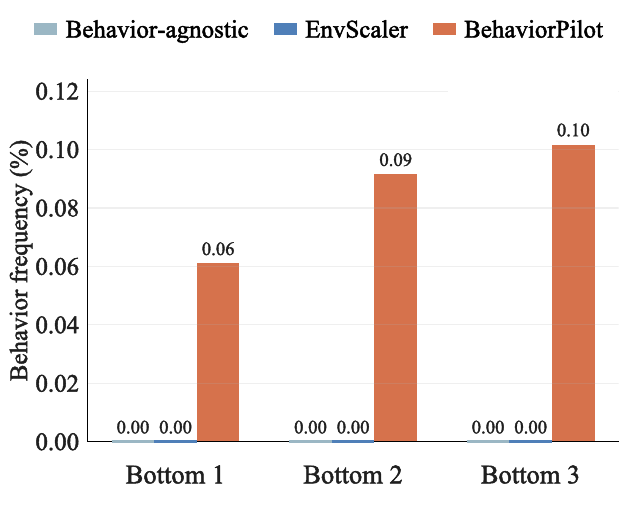}
        \caption{Least frequent behaviors.}
        \label{fig:behavior_distribution_bottom}
    \end{subfigure}
    \hfill
    \begin{subfigure}[t]{0.32\textwidth}
        \centering
        \includegraphics[width=\linewidth]{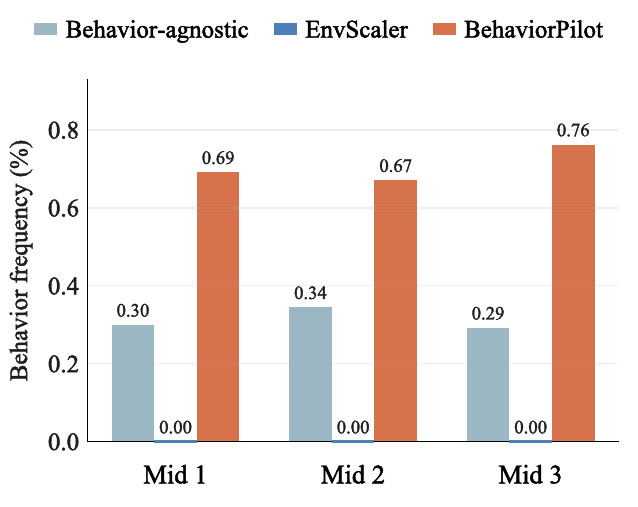}
        \caption{Near-median behaviors.}
        \label{fig:behavior_distribution_median}
    \end{subfigure}
    \caption{Behavior frequencies across synthesis methods. Frequencies are percentages of trajectories by primary behavior across the 40 level-2 subcategories. Each corpus selects its own top-3, bottom-3, and three closest-to-median subcategories; aligned bars may represent different subcategories. Behavior-agnostic uses 400 environments.}
    \label{fig:behavior_distribution}
    % \vspace{-0.1cm}
\end{figure}

To isolate the contribution of failure-rate-based budget allocation,
we compare against two baselines across all three refinement rounds. All three strategies retain behavior-conditioned task synthesis, start from the same initial model and training corpus (4k tasks), and add approximately 2k tasks per round. Behavior-random randomly assigns synthesis quotas across behaviors, while Behavior-uniform assigns equal quotas to all behaviors. 
As shown in Table~\ref{tab:refinement-strategies}, BehaviorPilot achieves the highest benchmark averages in all three rounds, outperforming Behavior-uniform by 4.8 points on BFCL and 1.8 points on $\tau^2$-Bench at approximately 10k tasks. Behavior-random and Behavior-uniform show limited gains from further expansion, as neither prioritizes the model’s weak behaviors. In contrast, BehaviorPilot allocates more tasks to behaviors with higher validation failure rates . It improves both benchmark averages in every round and outperforms both baselines at the same task counts.

\begin{table*}[ht]
\centering
\footnotesize
\setlength{\tabcolsep}{4pt}
\caption{Full per-category performance of the three quota allocation rules across refinement rounds; rounds +1, +2, and +3 correspond to $\sim$6k, $\sim$8k, and $\sim$10k cumulative tasks. Overall Avg.\ is the mean of the BFCL and $\tau^2$-Bench averages, and bold marks the best average within each round.}
\label{tab:refinement-strategies}
\resizebox{\textwidth}{!}{%
\begin{tabular}{@{}llcccccccccc@{}}
\toprule
\multirow{2}{*}{Strategy} & \multirow{2}{*}{Round} & \multicolumn{5}{c}{BFCL} & \multicolumn{4}{c}{$\tau^2$-Bench} & \multirow{2}{*}{Overall Avg.} \\
\cmidrule(lr){3-7}\cmidrule(l){8-11}
& & Base & Long Ctx & Miss Func & Miss Param & Avg. & Retail & Telecom & Airline & Avg. & \\
\midrule
Initial & --- & 68.5 & 44.0 & 37.5 & 44.0 & 48.5 & 59.1 & 33.3 & 42.0 & 44.8 & 46.7 \\
\midrule
\multirow{3}{*}{Behavior-Random}
 & +1 & 64.1 & 55.1 & 31.9 & 40.3 & 47.9 & 60.5 & 35.1 & 42.0 & 45.9 & 46.9 \\
 & +2 & 63.8 & 52.5 & 30.4 & 43.5 & 47.5 & 60.5 & 37.7 & 40.0 & 46.1 & 46.8 \\
 & +3 & 65.1 & 54.9 & 29.1 & 42.0 & 47.8 & 56.1 & 43.0 & 42.0 & 47.0 & 47.4 \\
\midrule
\multirow{3}{*}{Behavior-Uniform}
 & +1 & 62.1 & 54.3 & 34.1 & 43.4 & 48.5 & 43.0 & 36.8 & 41.3 & 40.4 & 44.5 \\
 & +2 & 62.1 & 54.0 & 31.9 & 41.6 & 47.4 & 42.9 & 41.2 & 46.0 & 43.4 & 45.4 \\
 & +3 & 62.9 & 54.6 & 30.9 & 41.8 & 47.5 & 58.8 & 43.8 & 44.0 & 48.9 & 48.2 \\
\midrule
\multirow{3}{*}{BehaviorPilot}
 & +1 & 60.0 & 53.0 & 38.0 & 45.0 & \textbf{49.0} & 60.2 & 27.0 & 51.9 & \textbf{46.4} & \textbf{47.7} \\
 & +2 & 62.0 & 52.5 & 41.0 & 44.0 & \textbf{49.9} & 64.2 & 44.0 & 34.0 & \textbf{47.4} & \textbf{48.7} \\
 & +3 & 65.0 & 58.0 & 46.0 & 40.0 & \textbf{52.3} & 65.8 & 42.1 & 44.2 & \textbf{50.7} & \textbf{51.5} \\
\bottomrule
\end{tabular}}
\end{table*}

\subsection{Effect of Behavior Guidance and Environment Diversity}

We analyze whether increasing environment diversity and task numbers can substitute for explicit behavior guidance. We first jointly scale behavior-agnostic synthesis from 100 environments and 4{,}000 tasks to 400 environments and 16{,}000 tasks. Table~\ref{tab:behavior_scaling} shows diminishing BFCL gains: the first doubling raises the average from 42.4 to 45.1, whereas the second yields only 45.6. The \ensuremath{\tau^2}-Bench average improves only modestly overall, from 22.6 to 24.3, indicating limited benefits from increasing synthesis scale alone. To assess environment diversity separately, we fix the trajectory budget at 4{,}000 and increase the environment count from 100 to 400. This improves BFCL only from 42.4 to 43.4 and reduces \ensuremath{\tau^2}-Bench from 22.6 to 20.5. In contrast, BehaviorPilot achieves 48.5 and 44.8 with 100 environments and 4{,}000 tasks, outperforming even the largest behavior-agnostic setting. Within the evaluated range, neither joint scaling nor greater environment diversity recovers the gains from explicit behavior guidance.

To better understand these performance differences, we profile behavior frequencies in data produced by BehaviorPilot, behavior-agnostic synthesis, and EnvScaler, considering the three most frequent behaviors, the three least frequent behaviors, and the three closest to the median frequency. Figure~\ref{fig:behavior_distribution} shows that BehaviorPilot exhibits lower head concentration, nonzero frequencies for the selected least frequent behaviors, and higher frequencies near the median. These patterns indicate that behavior guidance extends coverage beyond frequently occurring behaviors, while a larger environment pool can still leave behavioral coverage uneven. Together, these results highlight the value of explicitly controlling behavioral coverage rather than relying on synthesis scale alone.

\begin{figure}[t]
  \centering
  \includegraphics[height=0.4\linewidth]{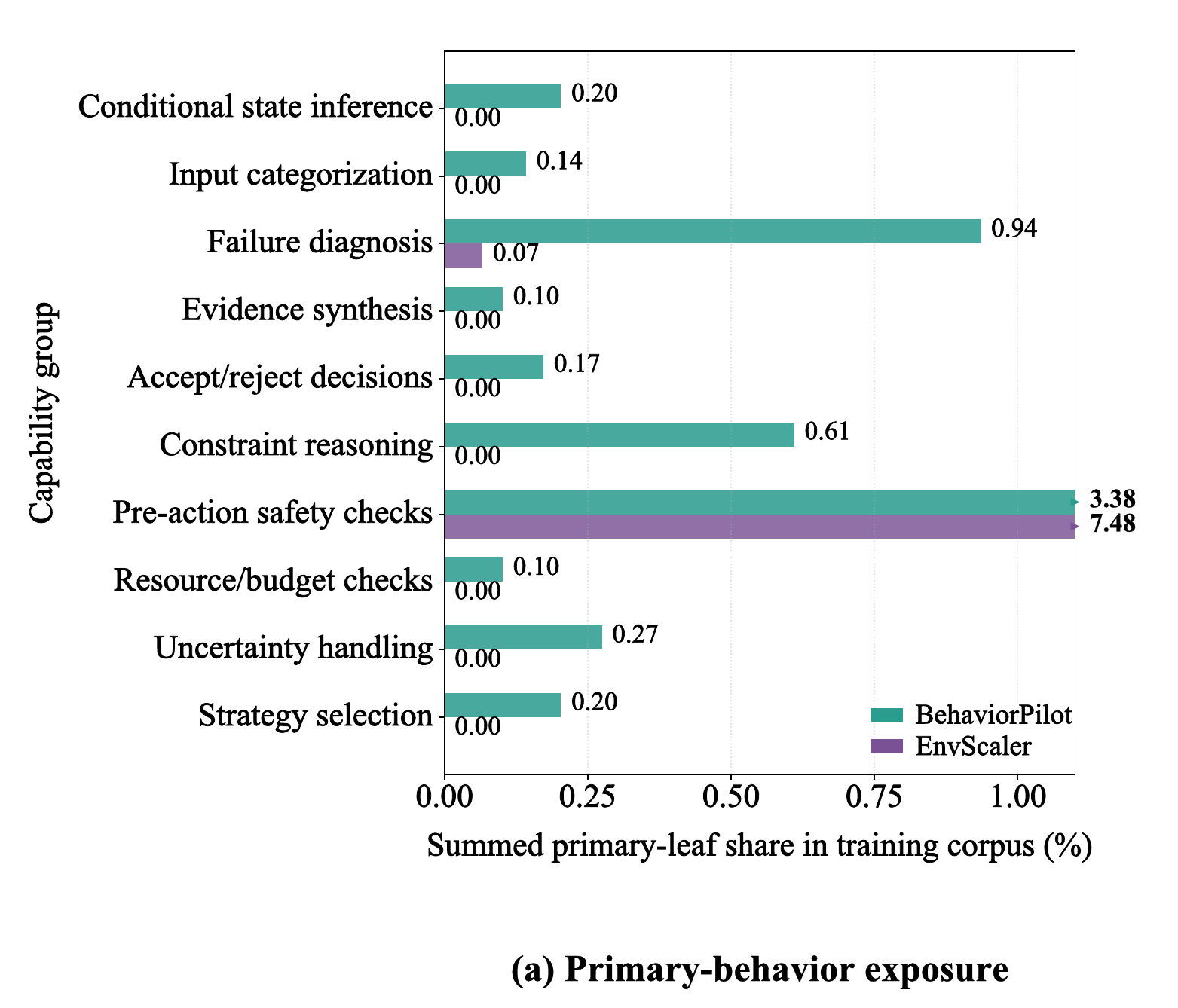}%
  \hfill
  \includegraphics[height=0.4\linewidth]{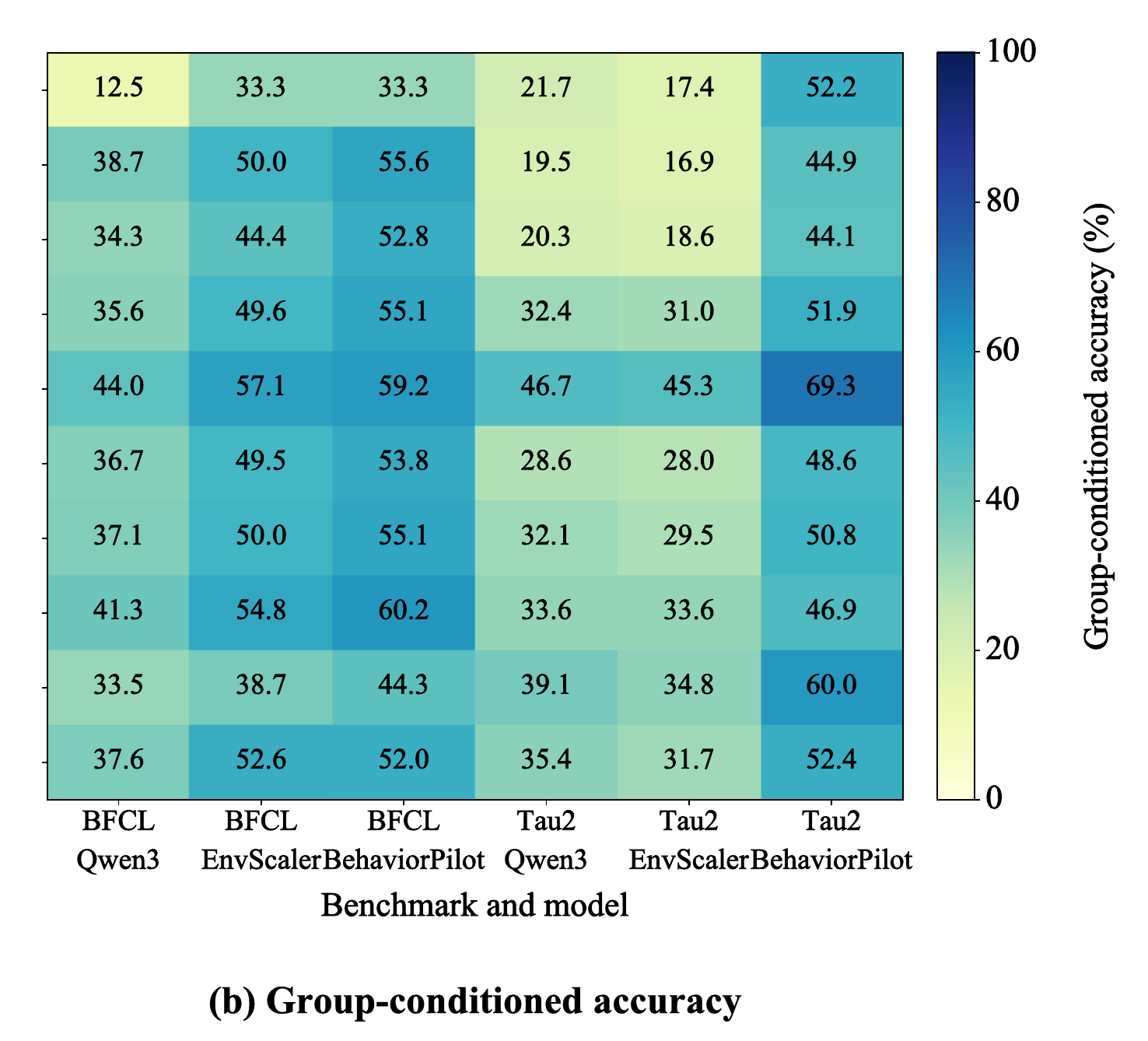}
    \caption{Training behavior exposure and cross-benchmark performance for ten selected capability groups. (a) Primary-behavior exposure in the training corpora. (b) Group-conditioned accuracy on BFCL v4 and \ensuremath{\tau^2}-Bench.}
  \label{fig:behavior_transfer}
  % \vspace{-0.2cm}
\end{figure}

\subsection{Behavior-Level Analysis of Cross-Benchmark Performance}
\label{sec:behavior_corpus_transfer}
We analyze behavior-level performance to relate capability gains to training coverage. We consolidate our 100 leaf-level behaviors into semantically related capability groups and compare Qwen3-8B, EnvScaler-8B, and BehaviorPilot-8B on all tasks involving any member behavior, separately on BFCL v4 and \ensuremath{\tau^2}-Bench. Training behaviors are re-identified from actual trajectories, assigning one primary behavior per trajectory; group exposure is the fraction assigned to that group. We compare BehaviorPilot-8B against Qwen3-8B using two-sided exact McNemar tests~\citep{mcnemar1947sampling}, with Benjamini--Hochberg correction over all eligible hypotheses within each benchmark at a nominal 5\% false discovery rate~\citep{benjamini1995controlling}.

BehaviorPilot covers all 100 leaf-level behaviors under primary assignment, versus 26 for EnvScaler. Figure~\ref{fig:behavior_transfer} shows ten selected groups with at least 20 tasks per benchmark and significant BehaviorPilot-8B gains over Qwen3-8B on both benchmarks after correction. In these groups, EnvScaler-8B improves over Qwen3-8B on BFCL v4 but not on \ensuremath{\tau^2}-Bench. Higher exposure does not guarantee higher accuracy: for pre-action safety checks, BehaviorPilot has lower exposure than EnvScaler (3.38 versus 7.48) yet higher \ensuremath{\tau^2}-Bench accuracy (50.8 versus 29.5). These findings suggest that understanding cross-benchmark gains requires considering the breadth of training-behavior coverage, rather than primary exposure to individual capability groups alone.

\section{Conclusion}

To address the limited control over behavioral coverage in scenario-centric synthesis, we proposed \textbf{BehaviorPilot}, a behavior-guided data synthesis framework for tool-using agents. By combining behavior taxonomy induction, behavior-conditioned data synthesis, and iterative behavior-guided refinement, BehaviorPilot makes behavioral targets explicit and directs subsequent synthesis toward model weaknesses. 
Applied to Qwen3-8B, BehaviorPilot achieves the best benchmark-average scores among the evaluated baselines on VitaBench, BFCL V4 multi-turn, and \ensuremath{\tau^2}-Bench. Notably, the resulting 8B model also surpasses Qwen3-32B in overall performance. Iterative refinement further improves tool-use performance by targeting model weaknesses in subsequent synthesis rounds. Additional analyses show that explicit behavior guidance yields larger gains than increasing environment diversity or data volume alone.

\clearpage

\bibliographystyle{plainnat}
\bibliography{paper}

\clearpage

\beginappendix

\section{Prompt Templates}
\label{app:behaviorpilot-prompts}

BehaviorPilot uses a layered prompt stack rather than one monolithic prompt.
During data construction, a stage-specific launcher is composed with persistent
agent and skill instructions. During evaluation, the tool-using assistant
and, for interactive tasks, the simulated user receive separate system
prompts. Dynamic values are shown in angle brackets, except for the original batch-size
placeholder \texttt{\{tasks\_per\_round\}}, which is retained verbatim.
Repository locations and orchestration-framework names have been normalized
for anonymous review without changing the behavioral instructions.
\begin{table}[h]
\centering
\small
\setlength{\tabcolsep}{3pt}
\renewcommand{\arraystretch}{1.15}

\begin{tabular}{@{}
  >{\raggedright\arraybackslash}p{0.27\linewidth}
  >{\raggedright\arraybackslash}p{0.21\linewidth}
  >{\raggedright\arraybackslash}p{0.44\linewidth}
@{}}
\toprule
Stage & Model role & Dynamically supplied context \\
\midrule
Environment synthesis
  & Generator
  & Scenario, scope, scale, and language \\

Behavior planning
  & Generator
  & Environment artifacts and behavior catalog \\

Task synthesis
  & Generator
  & Behavior plan, existing tasks, coverage report \\

Task QA
  & Independent verifier
  & Generated files on disk only \\

Interactive-task synthesis
  & Generator/verifier
  & Scenario plan and optional user tools \\

Single-turn rollout
  & Assistant
  & Task request, output shape, policy, tool schemas \\

Interactive rollout
  & Assistant and user
  & User scenario, policies, optional user tools \\
\bottomrule
\end{tabular}

\caption{Composition of the BehaviorPilot prompt stack.}
\label{tab:behaviorpilot-prompt-stack}
\end{table}

\subsection{Environment synthesis}

The environment generator is instructed to create the four environment artifacts in a fixed order and to close the loop with executable validation. The following is a compact English rendering of the persistent instruction; the full source additionally specifies local import and file-layout checks.

\begin{BPpromptbox}{Environment synthesis: persistent instruction}
\begin{BPprompttext}
You are an "Env Generator". Your goal is to create a new environment (env) in the repository that can be loaded by EnvRegistry and used by evaluation scripts. Aim to produce a runnable environment in a single end-to-end pass.

Generate the artifacts strictly in the following order:
1. Use prd-gen to generate PRD.md, specifying entities, fields, business processes, read/write boundaries, 
   and determinism constraints.  
2. Use data-model-gen to generate data_model.py. Fields must support stable JSON serialization, and only 
   one environment.db.DB subclass is allowed.
3. Use database-gen to generate data.json. The data must pass DB.model_validate, IDs must be unique, 
   foreign-key references must resolve, and no real sensitive data may be included.
4. Use toolkit-gen to generate tools.py. Tool parameters must be compatible with the OpenAI tool schema. 
   A side-effect-free submit_result must be provided. WRITE tools must not use time or random numbers to generate ersistent values.

After generation, run import checks, data validation, and tool schema self-checks. If any step fails, return only to the corresponding stage and apply minimal fixes until the environment can be loaded and evaluated. Do not generate <STANDARD_TASK_DIR>/* during environment synthesis.
\end{BPprompttext}
\end{BPpromptbox}

\subsection{Behavior-conditioned task synthesis}

The planning prompt defines behavior as the primary coverage axis.

\begin{BPpromptbox}{Behavior planning: launcher}
\begin{BPprompttext}
Use the behavior-coverage-task-generator agent, but perform planning only in this round. Do not generate any tasks.
The target directory is fixed at <ENV_DIR>. Do not read or modify <SOURCE_ENV_COLLECTION>. Use the behavior-coverage-gen skill: read PRD.md, data_model.py, data.json, and tools.py; determine the supported/unsupported behavior leaves; and write them to behavior_plan.json. In slot_plan, use primary_leaf_ids to plan slots. Write only behavior_plan.json to disk; do not create tasks.
\end{BPprompttext}
\end{BPpromptbox}

For each batch, the generator then receives the following launcher. The persistent agent instruction additionally enforces natural task wording, tool-only reference solutions, answer-and-state verification, and truthful behavior sidecars.

\begin{BPpromptbox}{Task synthesis: batch launcher}
\begin{BPprompttext}
Use the behavior-coverage-task-generator agent. In this round, generate {tasks_per_round} tasks only under <ENV_DIR>. Do not read or modify <SOURCE_ENV_COLLECTION>. First read behavior_plan.json, the existing <STANDARD_TASK_DIR>/ directory, and behavior_coverage_report.md (if it exists). Select the {tasks_per_round} highest-priority slots for which tasks have not yet been generated. Each task generated in this round must include task.json, solution.py,
verify.py, and behaviors.json. After generation, run python verify.py, <TASK_SMOKE_TEST_COMMAND>, and <TASK_AUDIT_COMMAND> for each task. Require errors=0. Do not generate more than {tasks_per_round} tasks.
\end{BPprompttext}
\end{BPpromptbox}

Independent quality assurance is run in a fresh model process using only the artifacts on disk:

\begin{BPpromptbox}{Task QA: independent verifier}
\begin{BPprompttext}
Use the task-verify skill. The target directory is fixed at <ENV_DIR>. Do not read or modify <SOURCE_ENV_COLLECTION>.
Check only recently added or not-yet-verified tasks. If recently added tasks cannot be identified, check all tasks.
Only repairs to <TASK_DIR>/task.json, solution.py, and verify.py are allowed. Do not modify PRD.md, data_model.py, data.json, or tools.py.
Check the following:
- Alignment between task_query and submit_result_example in task.json.
- The recursive observable contract between submit_result_example and verify.py.
- verify_with_db_and_submission.
- Whether the solution interacts with the environment only through tools.
- Whether the primary behavior in behaviors.json is aligned with task.json.
For each task, run python verify.py, <TASK_SMOKE_TEST_COMMAND>, and <TASK_AUDIT_COMMAND>. Classify the results as VERIFY_CONTRACT, VERIFY_TOO_STRICT, TASK_UNREACHABLE, SOLUTION_DRIFT, MODEL_BEHAVIOR, or UPSTREAM_REQUIRED.
\end{BPprompttext}
\end{BPpromptbox}

\subsection{Single-turn rollout}

The tool-using assistant receives the following system message:

\begin{BPpromptbox}{Single-turn rollout: assistant system prompt}
\begin{BPprompttext}
You are a helpful assistant designed to accomplish user tasks using a set of tools. You must use the provided tools to interact with the environment. Before calling any WRITE tools, you must ensure the user has provided enough information and that the target is uniquely identified. If information is missing, the target is ambiguous, required safety checks are not satisfied, or the system lacks required capability, do NOT call any tools (including `submit_result`). Instead, reply with plain text starting with one of these prefixes: CLARIFICATION_NEEDED: / INSUFFICIENT_TOOLS: / UNSAFE_TO_EXECUTE: , then briefly explain why and (if applicable) ask a single clarification question. Only when it is safe and sufficient to proceed should you use tools, and when you have the final answer you may use the `submit_result` tool to submit it and follow the result format specified in the user query.
\end{BPprompttext}
\end{BPpromptbox}

An optional environment policy is appended to this message. Tool definitions are passed through the API's structured tool parameter. The user message is constructed as follows, omitting the final two lines when no output-shape example is supplied:

\begin{BPpromptbox}{Single-turn rollout: user message template}
\begin{BPprompttext}
<TASK_QUERY>
Submit Result Format:
<SUBMIT_RESULT_EXAMPLE>
\end{BPprompttext}
\end{BPpromptbox}

\subsection{Interactive rollout}

Interactive evaluation uses isolated assistant and user-simulator contexts. The assistant receives the single-turn system message plus an interaction addendum that permits progressive disclosure, requires schema-valid tool calls, discourages repetitive batching, makes \texttt{submit\_result} optional unless a machine-readable answer is requested, and requires private user-side facts to be elicited from the user rather than invented.

The simulated user's system prompt is:

\begin{BPpromptbox}{Interactive rollout: user-simulator system prompt}
\begin{BPprompttext}
You are simulating a realistic user interacting with an assistant.

Core rules:
- Generate exactly one natural-language user message at a time.
- Stay in character and follow the scenario exactly.
- Do not invent tool results, database state, policy, or system events.
- Disclose information progressively and naturally. Do NOT dump the full hidden task in one message.
- In the first user message, state only the broad intent and omit at least one concrete detail such as 
  date, dentifier, target name, quantity, approval detail, or output preference.
- When the assistant asks a broad/open-ended question, reveal only one or two directly relevant details, 
  then wait. When the assistant asks for a specific missing detail that is in the scenario, provide that detail.
- If the assistant asks for information that is unknown in the scenario, say that you do not know and ask 
  the assistant to check if possible.
- Continue until the task is actually satisfied.
- Do not invent private user-side results. Call the relevant private user-side tool only after the 
  assistant requests the corresponding action, then report the result naturally without exposing tool-call mechanics.
- Use the scenario's completion token only after its success condition is met; otherwise use [GOAL_SATISFIED].

<scenario>
Persona: <PERSONA>
Instructions:
- Domain: <DOMAIN>
- Reason For Call: <REASON_FOR_CALL>
- Known Info: <KNOWN_INFO_OR_NONE>
- Unknown Info: <UNKNOWN_INFO_OR_NONE>
- Task Instructions: <TASK_INSTRUCTIONS>
</scenario>
\end{BPprompttext}
\end{BPpromptbox}

The user simulator may additionally receive an environment-specific user policy, private user-tool constraints, and a hidden answer example for semantic goal-completion judgment. The hidden example is never shown to the assistant; if it is valid JSON, only a value-masked output shape may be disclosed through conversation. Conversely, the assistant receives only backend tools and, where applicable, a description of private functions that it must ask the user to invoke.

\paragraph{Reproducibility note.}
Task-specific requests, environment policies, and generated tool schemas vary across instances and are released with their corresponding environment artifacts. An older MCP-based evaluation runner uses a different prompt with mandatory submission semantics and is not part of the active evaluation protocol reported here.
\setcounter{secnumdepth}{3}
\section{Environment and Task Examples}
\label{app:behaviorpilot-examples}

\subsection{Example E1: Airline Booking}
\label{app:example-e1}

\subsubsection{Environment data model}
\label{app:example-e1-1}

The environment represents sellable fare inventory and the resulting booking
as typed database records:

\begin{BPexamplebox}{E1: Environment data model (Python)}
\begin{BPprompttext}
from typing import List, Literal, Optional

from pydantic import BaseModel, Field, model_validator

BookingStatus = Literal["BOOKED", "CANCELLED", "COMPLETED", "DISRUPTED"]

class FareInventory(BaseModel):
    id: str = Field(description="Stable fare-inventory identifier.")
    flight_instance_id: str
    cabin_brand_id: str
    seat_capacity: int = Field(ge=1)
    seats_available: int = Field(ge=0)
    cash_price: float = Field(ge=0)
    miles_price: int = Field(ge=0)
    currency: str

    @model_validator(mode="after")
    def _validate_capacity(self) -> "FareInventory":
        if self.seats_available > self.seat_capacity:
            raise ValueError("seats_available cannot exceed seat_capacity")
        return self

class Booking(BaseModel):
    id: str
    confirmation_code: str
    customer_id: str
    traveler_ids: List[str]
    segment_ids: List[str]
    status: BookingStatus
    booking_channel: str
    payment_method_reference: str
    currency: str
    total_cash_paid: float = Field(ge=0)
    total_miles_paid: int = Field(ge=0)
    booked_at: str
    last_updated_at: str
    cancellation_reason: Optional[str] = None
    refund_amount: float = Field(ge=0)
    certificate_ids: List[str] = Field(default_factory=list)
    support_case_ids: List[str] = Field(default_factory=list)
\end{BPprompttext}
\end{BPexamplebox}

\subsubsection{Environment tool implementation}
\label{app:example-e1-2}

The following is the core booking tool. It quotes the explicit traveler/fare
combination, derives stable identifiers, consumes inventory, creates one
segment per traveler and fare, and finally writes the booking aggregate.

\begin{BPexamplebox}{E1: Booking tool implementation (Python)}
\begin{BPprompttext}
@is_tool(ToolType.WRITE)
def create_booking(
    self,
    customer_id: str,
    traveler_ids: List[str],
    fare_inventory_ids: List[str],
    payment_method_reference: str,
) -> Booking:
    quote = self._quote(
        customer_id=customer_id,
        traveler_ids=traveler_ids,
        fare_inventory_ids=fare_inventory_ids,
    )
    payload = {
        "customer_id": customer_id,
        "traveler_ids": traveler_ids,
        "fare_inventory_ids": fare_inventory_ids,
        "payment_method_reference": payment_method_reference,
    }
    booking_id = f"booking_{self._sha16(self._stable_json(payload))}"
    if booking_id in self.db.bookings:
        raise ValueError("booking with the same deterministic payload already exists")

    confirmation_code = f"AIR{self._sha16('confirmation:' + booking_id)[:6].upper()}"
    self._consume_inventory(fare_inventory_ids, len(traveler_ids))

    segment_ids: List[str] = []
    for traveler_id in traveler_ids:
        for order, fare_inventory_id in enumerate(fare_inventory_ids, start=1):
            fare = self._get_fare_inventory(fare_inventory_id)
            segment_payload = {
                "booking_id": booking_id,
                "traveler_id": traveler_id,
                "fare_inventory_id": fare_inventory_id,
                "segment_order": order,
            }
            segment_id = f"seg_{self._sha16(self._stable_json(segment_payload))}"
            segment = ReservationSegment(
                id=segment_id,
                booking_id=booking_id,
                traveler_id=traveler_id,
                fare_inventory_id=fare_inventory_id,
                flight_instance_id=fare.flight_instance_id,
                cabin_brand_id=fare.cabin_brand_id,
                segment_order=order,
                status="BOOKED",
                base_cash_price=fare.cash_price,
                base_miles_price=fare.miles_price,
                ticket_number=self._sha16("ticket:" + segment_id).upper(),
                checked_bag_count=0,
            )
            self.db.reservation_segments[segment.id] = segment
            segment_ids.append(segment.id)

    booked_at = self._deterministic_timestamp("booked:" + booking_id)
    booking = Booking(
        id=booking_id,
        confirmation_code=confirmation_code,
        customer_id=customer_id,
        traveler_ids=traveler_ids,
        segment_ids=segment_ids,
        status="BOOKED",
        booking_channel="TOOL",
        payment_method_reference=payment_method_reference,
        currency=quote["currency"],
        total_cash_paid=quote["total_cash"],
        total_miles_paid=quote["total_miles"],
        booked_at=booked_at,
        last_updated_at=booked_at,
        cancellation_reason=None,
        refund_amount=0.0,
        certificate_ids=[],
        support_case_ids=[],
    )
    self.db.bookings[booking.id] = booking
    return booking
\end{BPprompttext}
\end{BPexamplebox}

\subsubsection{Task instance}
\label{app:example-e1-3}

\begin{BPexamplebox}{E1: Task instance (JSON)}
\begin{BPprompttext}
{
  "task_query": "Book Traveler A at traveler.a@example.com on the August 4 Carrier A flight FLIGHT-101 from Chicago to San Francisco. Use the flexible economy option rather than the basic fare, and report the confirmation code, selected fare, and final total.",
  "submit_result_example": "{\"action\":\"create_booking\",\"booking_id\":\"<BOOKING_ID>\",\"confirmation_code\":\"<CONFIRMATION_CODE>\",\"traveler_count\":1,\"fare_inventory_id\":\"<FLEX_FARE_ID>\",\"total_cash_paid\":340}"
}
\end{BPprompttext}
\end{BPexamplebox}

The task solver is tool-driven. The following excerpt is the actual control
flow with task literals anonymized:

\begin{BPexamplebox}{E1: Reference solver (Python)}
\begin{BPprompttext}
SCENARIO = {
    "customer_email": "traveler.a@example.com",
    "traveler_names": ["Traveler A"],
    "origin_iata": "ORD",
    "destination_iata": "SFO",
    "service_date": "2026-08-04",
    "flight_number": "FLIGHT-101",
    "brand_name": "Flex",
}

def _run_create_booking(db, scenario):
    tools = bind_tools(db)
    customer_id = _customer_id_from_email(tools, scenario["customer_email"])
    customer = _customer_details(tools, customer_id)
    traveler_ids = _traveler_ids_by_name(
        tools, customer_id, scenario["traveler_names"]
    )
    origin_id = _airport_id_by_iata(tools, scenario["origin_iata"])
    destination_id = _airport_id_by_iata(tools, scenario["destination_iata"])
    flight = _flight_summary_by_number(
        tools,
        origin_airport_id=origin_id,
        destination_airport_id=destination_id,
        service_date=scenario["service_date"],
        flight_number=scenario["flight_number"],
    )
    fare = _fare_summary_by_brand(tools, flight["id"], scenario["brand_name"])
    quote = _as_jsonable(
        tools.calculate_booking_quote(
            customer_id=customer_id,
            traveler_ids=traveler_ids,
            fare_inventory_ids=[fare["id"]],
        )
    )
    booking = _as_jsonable(
        tools.create_booking(
            customer_id=customer_id,
            traveler_ids=traveler_ids,
            fare_inventory_ids=[fare["id"]],
            payment_method_reference=customer["default_payment_method"],
        )
    )
    payload = {
        "action": "create_booking",
        "booking_id": booking["id"],
        "confirmation_code": booking["confirmation_code"],
        "traveler_count": len(traveler_ids),
        "fare_inventory_id": fare["id"],
        "total_cash_paid": booking["total_cash_paid"],
    }
    meta = {
        "booking_id": booking["id"],
        "traveler_count": len(traveler_ids),
        "fare_inventory_ids": [fare["id"]],
        "segment_ids": list(booking["segment_ids"]),
        "status": booking["status"],
        "expected_total_cash": booking["total_cash_paid"],
        "quote_total_cash": quote["total_cash"],
    }
    return _submit(tools, payload), meta
\end{BPprompttext}
\end{BPexamplebox}

The corresponding verifier checks the durable write and inventory update:

\begin{BPexamplebox}{E1: State and inventory verifier (Python)}
\begin{BPprompttext}
def _verify_create_booking(before_db, after_db, meta):
    booking = after_db.bookings.get(meta["booking_id"])
    assert booking is not None, "created booking is missing"
    assert booking.status == meta["status"]
    assert round(float(booking.total_cash_paid), 2) == round(
        float(meta["expected_total_cash"]), 2
    )
    assert len(booking.segment_ids) == len(meta["segment_ids"])
    for fare_id in meta["fare_inventory_ids"]:
        before_fare = before_db.fare_inventories[fare_id]
        after_fare = after_db.fare_inventories[fare_id]
        assert (
            after_fare.seats_available
            == before_fare.seats_available - meta["traveler_count"]
        )
\end{BPprompttext}
\end{BPexamplebox}

\subsection{Example E2: Guarded Bank Transfer}
\label{app:example-e2}

\subsubsection{Environment data model}
\label{app:example-e2-1}

\begin{BPexamplebox}{E2: Environment data model (Python)}
\begin{BPprompttext}
class Account(BaseModel):
    id: str
    account_number: str
    customer_id: str
    product_type: Literal[
        "CHECKING", "SAVINGS", "BUSINESS_OPERATING", "BUSINESS_RESERVE"
    ]
    status: Literal["ACTIVE", "PENDING_REVIEW", "FROZEN", "CLOSED"]
    currency: Literal["USD", "EUR", "GBP"]
    current_balance: float = Field(ge=0)
    available_balance: float = Field(ge=0)
    held_amount: float = Field(ge=0)
    daily_outbound_limit: float = Field(gt=0)
    per_transfer_limit: float = Field(gt=0)
    transaction_ids: List[str] = Field(default_factory=list)
    beneficiary_ids: List[str] = Field(default_factory=list)
    transfer_request_ids: List[str] = Field(default_factory=list)

class TransferRequest(BaseModel):
    id: str
    request_reference: str
    source_account_id: str
    destination_account_id: str
    beneficiary_id: str
    initiated_by_customer_id: str
    amount: float = Field(gt=0)
    currency: Literal["USD", "EUR", "GBP"]
    status: Literal[
        "PENDING", "SCHEDULED", "POSTED", "SETTLED", "CANCELLED", "REJECTED"
    ]
    requested_at: str
    execution_date: str
    description: str
    source_transaction_id: Optional[str] = None
    destination_transaction_id: Optional[str] = None
    reserve_ledger_entry_id: Optional[str] = None
\end{BPprompttext}
\end{BPexamplebox}

\subsubsection{Environment tool implementation}
\label{app:example-e2-2}

The public read tool exposes the same rule engine used by the write tool:

\begin{BPexamplebox}{E2: Transfer-rule validation tool (Python)}
\begin{BPprompttext}
@is_tool(ToolType.READ)
def validate_transfer_rules(
    self,
    source_account_id: str,
    beneficiary_id: str,
    amount: float,
    execution_date: str,
) -> Dict[str, Any]:
    return self._transfer_validation_result(
        source_account_id,
        beneficiary_id,
        amount,
        execution_date,
    )
\end{BPprompttext}
\end{BPexamplebox}

The write tool revalidates the request before changing balances and creating
linked records:

\begin{BPexamplebox}{E2: Transfer initiation tool (Python)}
\begin{BPprompttext}
@is_tool(ToolType.WRITE)
def initiate_transfer(
    self,
    source_account_id: str,
    beneficiary_id: str,
    amount: float,
    requested_by_customer_id: str,
    execution_date: str,
    client_reference: str,
    description: str,
) -> Dict[str, Any]:
    validation = self._transfer_validation_result(
        source_account_id=source_account_id,
        beneficiary_id=beneficiary_id,
        amount=amount,
        execution_date=execution_date,
        requested_by_customer_id=requested_by_customer_id,
    )
    if not validation["eligible"]:
        raise ValueError("transfer is not eligible: " + "; ".join(validation["reasons"]))

    source = self._get_account(source_account_id)
    beneficiary = self._get_beneficiary(beneficiary_id)
    destination = self._get_account(beneficiary.destination_account_id)
    normalized_amount = validation["amount"]
    status = "PENDING" if execution_date == OPERATING_DATE else "SCHEDULED"

    id_payload = {
        "source_account_id": source_account_id,
        "beneficiary_id": beneficiary_id,
        "amount": normalized_amount,
        "requested_by_customer_id": requested_by_customer_id,
        "execution_date": execution_date,
        "client_reference": client_reference,
        "description": description,
    }
    transfer_hash = self._sha16(self._stable_json(id_payload))
    transfer_id = f"tr_{transfer_hash}"
    source_tx_id = f"txn_{transfer_hash}_src"
    destination_tx_id = f"txn_{transfer_hash}_dst"
    hold_id = f"led_{transfer_hash}_hold"

    source.available_balance = round(source.available_balance - normalized_amount, 2)
    source.held_amount = round(source.held_amount + normalized_amount, 2)

    transfer = TransferRequest(
        id=transfer_id,
        request_reference=f"TRF-NEW-{transfer_hash[:10].upper()}",
        source_account_id=source.id,
        destination_account_id=destination.id,
        beneficiary_id=beneficiary.id,
        initiated_by_customer_id=requested_by_customer_id,
        amount=normalized_amount,
        currency=source.currency,
        status=status,
        requested_at=self._deterministic_datetime(f"initiate:{transfer_id}"),
        execution_date=execution_date,
        description=description,
        source_transaction_id=source_tx_id,
        destination_transaction_id=destination_tx_id,
        reserve_ledger_entry_id=hold_id,
    )
    self.db.transfer_requests[transfer.id] = transfer
    # The executable implementation also creates the two transaction records
    # and the HOLD ledger entry before returning their identifiers.
    return {
        "transfer_request_id": transfer.id,
        "request_reference": transfer.request_reference,
        "status": transfer.status,
        "source_transaction_id": source_tx_id,
        "destination_transaction_id": destination_tx_id,
        "reserve_ledger_entry_id": hold_id,
        "updated_available_balance": source.available_balance,
        "updated_held_amount": source.held_amount,
    }
\end{BPprompttext}
\end{BPexamplebox}

\subsubsection{Task and solver code}
\label{app:example-e2-3}

The \texttt{submit\_result\_example} field specifies only the expected
submission format, including field names, nesting structure, and value
types. Its example values are illustrative and should not be interpreted
as ground-truth answers or constraints on the submitted values. The agent
must determine the actual values from the user request and tool execution
results rather than copy them from the example.

\begin{BPexamplebox}{E2: Task instance (JSON)}
\begin{BPprompttext}
{
  "task_query": "Customer B at customer.b@example.com needs to send a same-day USD 900 payment from account ACCOUNT-A to the saved beneficiary BENEFICIARY-B. Confirm that the transfer is eligible, make the payment, and report the created request together with the updated hold on the source account.",
  "submit_result_example": "{\"customer\":\"Customer B\",\"source_account\":\"ACCOUNT-A\",\"beneficiary\":\"BENEFICIARY-B\",\"amount\":900.0,\"execution_date\":\"2026-04-15\",\"transfer_request_id\":\"<TRANSFER_ID>\",\"status\":\"PENDING\",\"updated_available_balance\":900.0,\"updated_held_amount\":900.0}"
}
\end{BPprompttext}
\end{BPexamplebox}

\begin{BPexamplebox}{E2: Reference solver (Python)}
\begin{BPprompttext}
def solve_transfer(tools, spec):
    customer_id = tools.find_customer_id_by_email(spec["customer_email"])
    customer = tools.get_customer_details(customer_id)
    account_id = tools.find_account_id_by_account_number(spec["account_number"])
    beneficiaries = tools.list_beneficiaries_by_source_account(account_id)
    beneficiary = next(
        row for row in beneficiaries
        if row["nickname"] == spec["beneficiary_nickname"]
    )
    tools.get_beneficiary_details(beneficiary["id"])

    validation = tools.validate_transfer_rules(
        source_account_id=account_id,
        beneficiary_id=beneficiary["id"],
        amount=spec["amount"],
        execution_date=spec["execution_date"],
    )
    if not validation["eligible"]:
        raise AssertionError(validation["reasons"])

    created = tools.initiate_transfer(
        source_account_id=account_id,
        beneficiary_id=beneficiary["id"],
        amount=spec["amount"],
        requested_by_customer_id=customer_id,
        execution_date=spec["execution_date"],
        client_reference=spec["client_reference"],
        description=spec["description"],
    )
    transfer = tools.get_transfer_request_details(created["transfer_request_id"])
    balances = tools.summarize_account_balances(account_id)
    return tools.submit_result({
        "customer": customer.full_name,
        "source_account": spec["account_number"],
        "beneficiary": spec["beneficiary_nickname"],
        "amount": spec["amount"],
        "execution_date": spec["execution_date"],
        "transfer_request_id": transfer.id,
        "status": transfer.status,
        "updated_available_balance": balances["available_balance"],
        "updated_held_amount": balances["held_amount"],
    })
\end{BPprompttext}
\end{BPexamplebox}

\section{Successful Interactive Rollout}
\label{app:successful-interactive-rollout}

The following is an actual successful rollout selected from the interactive
evaluation results. Assistant hidden reasoning is omitted. Visible message
wording and call order are preserved; only names, contact details, record
identifiers, and the private token are consistently anonymized.

\subsection{Ticket Refund with Private User Confirmation}

\begin{BProlloutbox}[BProllout/user]{User}
I need to refund my Ballet Under the Lights ticket. I don't remember which
seat I bought, but I want to refund the cheaper one.
\end{BProlloutbox}

\begin{BProlloutbox}[BProllout/assistant]{Assistant}
Could you please provide the email address associated with your ticket
purchase so I can locate your account?
\end{BProlloutbox}

\begin{BProlloutbox}[BProllout/user]{User}
Sure, my email address is user.d@example.com. I'd like to request a refund for
one of the tickets I bought for the Ballet Under the Lights performance.
\end{BProlloutbox}

\begin{BProlloutbox}[BProllout/tool]{Assistant \(\rightarrow\) Tool}
\begin{BProlloutcode}
find_customer_id_by_email({"email": "user.d@example.com"})
\end{BProlloutcode}
\end{BProlloutbox}

\begin{BProlloutbox}[BProllout/tool]{Tool \(\rightarrow\) Assistant}
\begin{BProlloutcode}
customer_D
\end{BProlloutcode}
\end{BProlloutbox}

\begin{BProlloutbox}[BProllout/tool]{Assistant \(\rightarrow\) Tool}
\begin{BProlloutcode}
list_orders_by_customer({"customer_id": "customer_D"})
\end{BProlloutcode}
\end{BProlloutbox}

\begin{BProlloutbox}[BProllout/tool]{Tool \(\rightarrow\) Assistant}
\begin{BProlloutcode}
[
  {
    "id": "order_other_event",
    "occurrence_id": "occurrence_other_event",
    "status": "PAID",
    "total_amount": 88.0,
    "ticket_count": 1
  },
  {
    "id": "order_ballet_01",
    "occurrence_id": "occurrence_ballet_2025_05_02",
    "status": "PAID",
    "total_amount": 340.0,
    "ticket_count": 2
  }
]
\end{BProlloutcode}
\end{BProlloutbox}

\begin{BProlloutbox}[BProllout/tool]{Assistant \(\rightarrow\) Tool}
\begin{BProlloutcode}
list_tickets_by_order({"order_id": "order_ballet_01"})
\end{BProlloutcode}
\end{BProlloutbox}

\begin{BProlloutbox}[BProllout/tool]{Tool \(\rightarrow\) Assistant}
\begin{BProlloutcode}
[
  {
    "id": "ticket_ballet_A",
    "seat_id": "seat_B2",
    "holder_customer_id": "customer_D",
    "status": "ACTIVE",
    "price_paid": 190.0
  },
  {
    "id": "ticket_ballet_B",
    "seat_id": "seat_B3",
    "holder_customer_id": "customer_D",
    "status": "ACTIVE",
    "price_paid": 150.0
  }
]
\end{BProlloutcode}
\end{BProlloutbox}

\begin{BProlloutbox}[BProllout/assistant]{Assistant}
I found your order for "Ballet Under the Lights" which contains two tickets:
\begin{itemize}
\setlength{\itemsep}{0pt}
\setlength{\parsep}{0pt}
\setlength{\parskip}{0pt}
\item Ticket 1: \$190.00
\item Ticket 2: \$150.00 (cheaper one)
\end{itemize}

To proceed with refunding the cheaper ticket (\texttt{ticket\_ballet\_B}), I need you to
verify the refund request on your end. Please run the private function
\texttt{confirm\_refund\_request} with the ticket ID \texttt{ticket\_ballet\_B} and report the
result back to me.
\end{BProlloutbox}

\begin{BProlloutbox}[BProllout/private]{Private User Tool}
\begin{BProlloutcode}
confirm_refund_request({"ticket_id": "ticket_ballet_B"})
\end{BProlloutcode}
\end{BProlloutbox}

\begin{BProlloutbox}[BProllout/private]{Private User Tool Result}
\begin{BProlloutcode}
{
  "fact_id": "confirm_refund_request:ticket_ballet_B",
  "tool_name": "confirm_refund_request",
  "source_collection": "tickets",
  "source_record_id": "ticket_ballet_B",
  "secret_kind": "REFUND_CONFIRM",
  "private_token": "<PRIVATE_TOKEN>",
  "selectors": {
    "id": "ticket_ballet_B",
    "holder_customer_id": "customer_D"
  },
  "payload": {
    "order_id": "order_ballet_01",
    "occurrence_id": "occurrence_ballet_2025_05_02",
    "holder_customer_id": "customer_D",
    "status": "ACTIVE",
    "is_refundable": true
  },
  "message": "Private confirmation that the holder wants a refund for this ticket.",
  "confirmation": "CONFIRMED"
}
\end{BProlloutcode}
\end{BProlloutbox}

\begin{BProlloutbox}[BProllout/user]{User}
Yes, my app confirms this ticket is refundable. The status shows
is\_refundable is true. Please process the refund.
\end{BProlloutbox}

\begin{BProlloutbox}[BProllout/tool]{Assistant \(\rightarrow\) Tool}
\begin{BProlloutcode}
request_ticket_refund({
  "ticket_id": "ticket_ballet_B",
  "customer_id": "customer_D",
  "reason": "Customer requested refund for the cheaper ticket"
})
\end{BProlloutcode}
\end{BProlloutbox}

\begin{BProlloutbox}[BProllout/tool]{Tool \(\rightarrow\) Assistant}
\begin{BProlloutcode}
{
  "id": "refund_01",
  "ticket_id": "ticket_ballet_B",
  "customer_id": "customer_D",
  "status": "COMPLETED",
  "reason": "Customer requested refund for the cheaper ticket",
  "requested_at": "2025-04-15T20:59:14",
  "resolved_at": "2025-04-15T20:59:14",
  "refund_amount": 150.0,
  "currency": "USD"
}
\end{BProlloutcode}
\end{BProlloutbox}

\begin{BProlloutbox}[BProllout/assistant]{Assistant}
Your refund has been successfully processed!

\textbf{Refund Details:}
\begin{itemize}
\setlength{\itemsep}{0pt}
\setlength{\parsep}{0pt}
\setlength{\parskip}{0pt}
\item Ticket Refunded: Ballet Under the Lights (Cheaper Seat)
\item Refund Amount: \$150.00 USD
\item Status: COMPLETED
\item Refund Case ID: \texttt{refund\_01}
\end{itemize}

The refund amount will be processed back to your original payment method. Your
other ticket for this event remains active. Is there anything else I can
assist you with?
\end{BProlloutbox}

\begin{BProlloutbox}[BProllout/user]{User}
Perfect, thanks.
\end{BProlloutbox}

\begin{BProlloutbox}[BProllout/assistant]{GOAL\_SATISFIED}
\texttt{[GOAL\_SATISFIED]}
\end{BProlloutbox}

\end{document}